%% file: arxiv.tex
\documentclass{article}

\PassOptionsToPackage{numbers, sort&compress}{natbib}

\usepackage[preprint]{arxiv}

\usepackage[utf8]{inputenc}
\usepackage[T1]{fontenc}
\usepackage[pagebackref,breaklinks,colorlinks]{hyperref}
\usepackage{url}
\usepackage{booktabs}
\usepackage{amsfonts}
\usepackage{nicefrac}
\usepackage{microtype}
\usepackage[table]{xcolor}

\usepackage[pdftex]{graphicx}
\usepackage{amsmath}
\usepackage{amssymb}
\usepackage{multirow}
\usepackage{footnote}
\usepackage{multibib}
\usepackage{enumitem}
\usepackage{makecell}
\usepackage{graphbox}
\usepackage{float}
\usepackage{caption}
\usepackage{tcolorbox}
\tcbuselibrary{skins, breakable}
\usepackage{subfigure}
\newcites{appendix}{Appendix References}

\usepackage{authblk}

\usepackage[misc]{ifsym}

\usepackage[capitalize]{cleveref}
\crefname{section}{Sec.}{Secs.}
\Crefname{section}{Section}{Sections}
\Crefname{table}{Table}{Tables}
\crefname{table}{Tab.}{Tabs.}

\def\ie{\emph{i.e.}}
\def\eg{\emph{e.g.}}

\newcommand{\mccraftingtable}{\includegraphics[width=0.025\paperwidth,align=c]{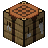}}
\newcommand{\mcwoodenpickaxe}{\includegraphics[width=0.025\paperwidth,align=c]{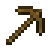}}
\newcommand{\mcstonepickaxe}{\includegraphics[width=0.025\paperwidth,align=c]{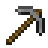}}
\newcommand{\mcironpickaxe}{\includegraphics[width=0.025\paperwidth,align=c]{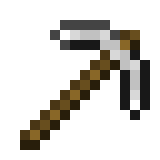}}
\newcommand{\mcdiamond}{\includegraphics[width=0.025\paperwidth,align=c]{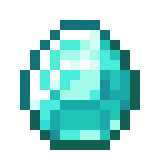}}

\title{Ghost in the Minecraft: Generally Capable Agents for Open-World Environments via Large Language Models with Text-based Knowledge and Memory}

\author[ ]{Xizhou Zhu$^{1,2}$\thanks{Equal contribution. This work is done when Chenxin Tao and Weijie Su are interns at SenseTime Research. \Letter\ Corresponding to Jifeng Dai <daijifeng@tsinghua.edu.cn>.}}
\author[ ]{Yuntao Chen$^{3*}$}
\author[ ]{Hao Tian$^{2*}$}
\author[ ]{Chenxin Tao$^{1,2*}$}
\author[ ]{Weijie Su$^{2,4*}$}
\author[ ]{Chenyu Yang$^{1*}$}
\author[1]{\authorcr Gao Huang}
\author[4]{Bin Li}
\author[2]{Lewei Lu}
\author[2,5]{Xiaogang Wang}
\author[6]{Yu Qiao}
\author[7]{Zhaoxiang Zhang}
\author[1,6~\Letter]{Jifeng Dai}
\affil[1]{Tsinghua University \quad $^2$SenseTime Research}
\affil[3]{Centre for Artificial Intelligence and Robotics, HKISI, CAS }
\affil[4]{University of Science and Technology of China}
\affil[5]{The Chinese University of Hong Kong \quad $^6$Shanghai Artificial Intelligence Laboratory}
\affil[7]{Institute of Automation, Chinese Academy of Science (CASIA)}
\affil[ ]{\tt\small \{zhuxizhou,gaohuang,daijifeng\}@tsinghua.edu.cn, chenyuntao08@gmail.com}
\affil[ ]{\tt\small tianhao2@senseauto.com, \{tcx20,yangcy19\}@mails.tsinghua.edu.cn, jackroos@mail.ustc.edu.cn, binli@ustc.edu.cn, luotto@sensetime.com}
\affil[ ]{\tt\small xgwang@ee.cuhk.edu.hk, qiaoyu@pjlab.org.cn, zhaoxiang.zhang@ia.ac.cn}

\begin{document}

\maketitle

\vspace{-1.5em}
\input{srcs/01_abstract}

\vspace{-1.0em}
\input{srcs/02_introduction}

\input{srcs/03_relatedwork}

\input{srcs/04_method}

\input{srcs/05_experiments}

\input{srcs/06_conclusion}

\paragraph{Acknowledgments} The work is partially supported by the National Natural Science Foundation of China under grants No.U19B2044, No.61836011, No.62022048, and No.62276150. This work is also partially supported by the National Key R\&D Program of China under grants NO.2022ZD0114900, and the Guoqiang Institute of Tsinghua University.

{\small
\bibliographystyle{abbrvnat}
\bibliography{egbib}
}

\appendix
\input{srcs/07_appendix}

\end{document}

%% file: srcs/01_abstract.tex
\begin{abstract}
The captivating realm of Minecraft has attracted substantial research interest in recent years, serving as a rich platform for developing intelligent agents capable of functioning in open-world environments. However, the current research landscape predominantly focuses on specific objectives, such as the popular "ObtainDiamond" task, and has not yet shown effective generalization to a broader spectrum of tasks. 
Furthermore, the current leading success rate for the "ObtainDiamond" task stands at around 20\%, highlighting the limitations of Reinforcement Learning (RL) based controllers used in existing methods.
To tackle these challenges, we introduce Ghost in the Minecraft (GITM), a novel framework integrates Large Language Models (LLMs) with text-based knowledge and memory, aiming to create Generally Capable Agents (GCAs) in Minecraft. These agents, equipped with the logic and common sense capabilities of LLMs, can skillfully navigate complex, sparse-reward environments with text-based interactions.
We develop a set of structured actions and leverage LLMs to generate action plans for the agents to execute. 
The resulting LLM-based agent markedly surpasses previous methods, achieving a remarkable improvement of +47.5\% in success rate on the "ObtainDiamond" task, demonstrating superior robustness compared to traditional RL-based controllers.
Notably, our agent is the first to procure all items in the Minecraft Overworld technology tree, demonstrating its extensive capabilities. GITM does not need any GPU for training, but a single CPU node with 32 CPU cores is enough. This research shows the potential of LLMs in developing capable agents for handling long-horizon, complex tasks and adapting to uncertainties in open-world environments. See the project website at \url{https://github.com/OpenGVLab/GITM}.
\end{abstract}

%% file: srcs/02_introduction.tex
\section{Introduction}

\emph{“What if a cyber brain could possibly generate its own ghost, create a soul all by itself? And if it did, just what would be the importance of being human then?”}

\rightline{--- Ghost in the Shell (1995) }

Minecraft, as the world's best-selling game, boasts over 238 million copies sold and more than 140 million peak monthly active users~\cite{enwiki:1155148900}. 
Within the game, hundreds of millions of players have experienced a digital second life by surviving, exploring and creating, closely resembling the human world in many aspects. 
Given its massive scale, vast success, and unrestricted freedom, Minecraft has established itself as an unparalleled platform for researching autonomous and robust \textit{Generally Capable Agents (GCAs)}~\cite{team2021open} in open-world environments brimmed with long-horizon challenges, environmental disruptions, and uncertainties.

\begin{figure}[t]
  \small
  \centering
  \includegraphics[width=1.0\linewidth]{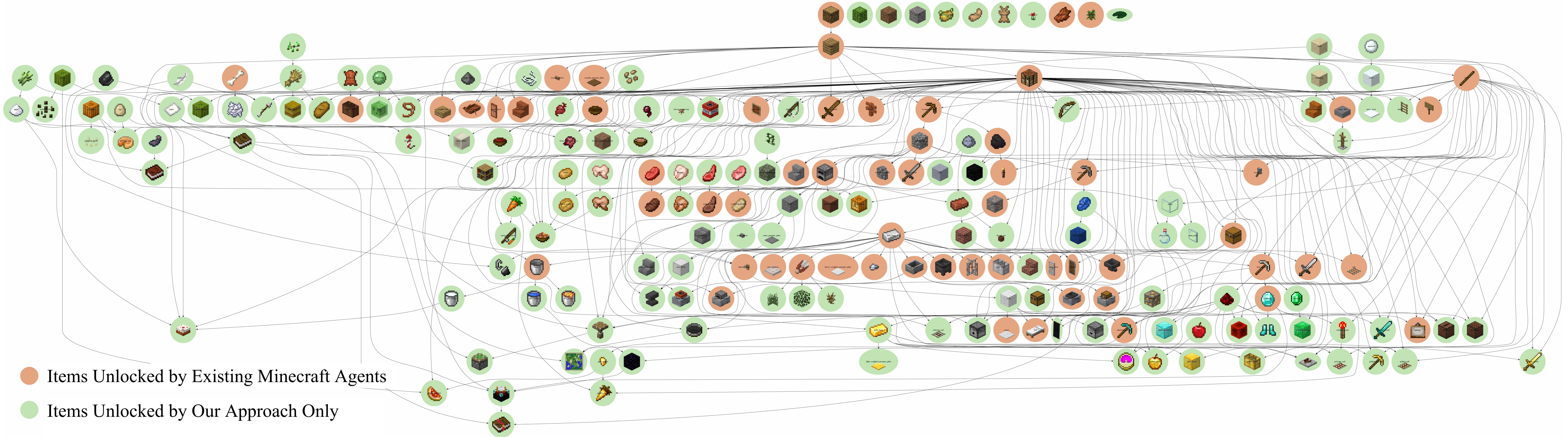}
  \caption{\textbf{Our GITM unlocks the entire technology tree by obtaining all items in Minecraft Overworld.} Each node represents an individual item in Minecraft. The directed edges between nodes represent prerequisite relationships for obtaining items. For better readability, we manually merge some similar nodes, \eg, ``wooden\_pickaxe'', ``wooden\_axe'', ``wooden\_hoe'', and 'wooden\_shovel' are merged into one node, and ``wooden\_pickaxe'' is selected to represent the merged node. Existing Minecraft agents~\cite{baker2022video,hafner2023mastering,wang2023describe} only unlocked 78 / 262 = 30\% items, while our GITM successfully unlocked all items. }
  \label{fig:item_graph}
  
\end{figure}

Minecraft acts as a microcosm of the real world. Developing an automated agent that can master all technical challenges in Minecraft is akin to creating an artificial intelligence capable of autonomously learning and mastering the entire real-world technology.
However, existing researches~\cite{wang2023describe,baker2022video,hafner2023mastering} remain narrowly scoped. Prior studies have predominantly focused on the specific goal of \texttt{ObtainDiamond}~\cite{milani2023towards}. Yet, in the process of obtaining diamonds, the number of types of items involved only accounts for $<$5\% of the entire Minecraft world. \texttt{ObtainDiamond} only requires specialized skills in a specific domain, while obtaining all items in Minecraft demonstrates a wide range of knowledge and capabilities, similar to mastering multidisciplinary fields in the real world.
As illustrated in Fig.~\ref{fig:item_graph}, our work endeavors to obtain all items in Minecraft within a reasonable computation budget. This achievement stands as a significant milestone in the development of GCAs, illustrating the potential of intelligent agents to match human performance in terms of versatility and adaptability.

\begin{figure}[t]
  \small
  \centering  \includegraphics[width=0.9\linewidth]{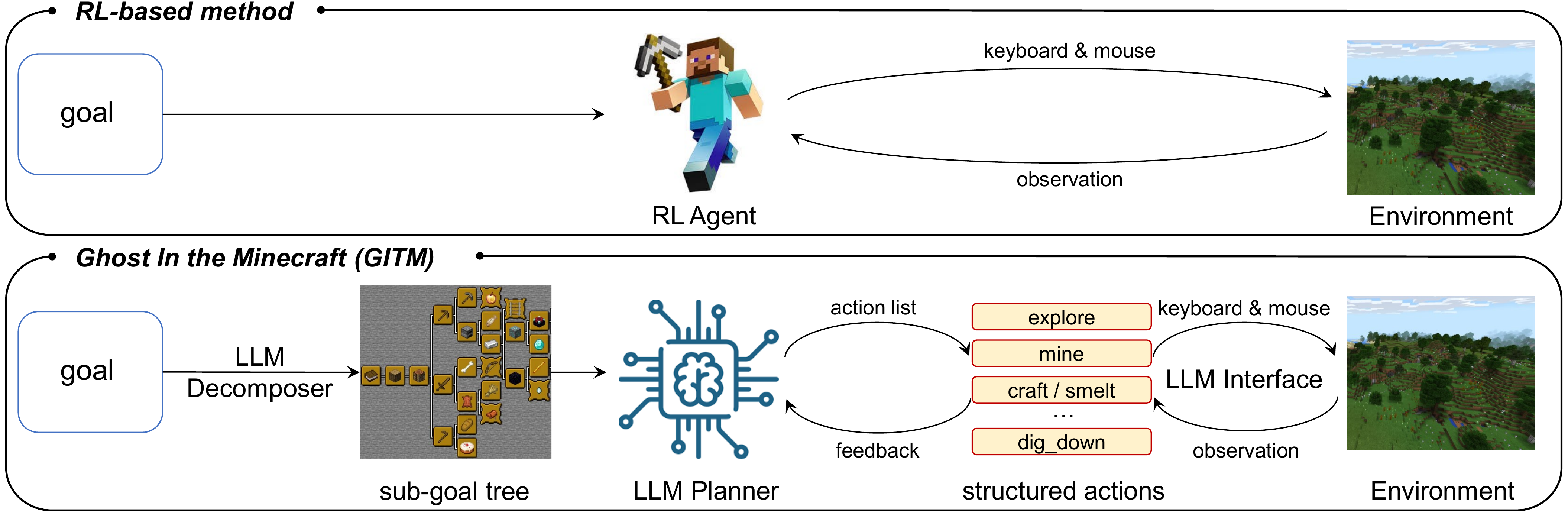}
  \vspace{-0.5em}
  \caption{\textbf{Comparison between RL-based method and our GITM.} RL agents try to map an complex goal directly to a sequence of low-level control signals, while our GITM leverages LLM to break down the goals and map them to structured actions for final control signals.}
  \label{fig:comparison}
  \vspace{-1em}
\end{figure}

Although reinforcement learning (RL)~\cite{matsuo2022deep} is the most popular paradigm for approaching GCAs, it has shown some staggering limitations in conquering Minecraft.
RL-based agents typically require a vast number of learning steps (\eg, nearly 30 million steps to obtain diamonds as reported in DreamerV3~\cite{hafner2023mastering}) and exhibit poor scalability when learning new tasks(\eg, VPT~\cite{baker2022video} uses different agents for world exploration and diamond mining). 
As a consequence, adopting RL-based agents for completing a wide range of tasks may require an prohibitively high number of training steps, making it impractical to obtain all items in Minecraft. 
This inefficiency and lack of adaptability have hindered the development of generally capable agents in open-world environments.

As shown in Fig.~\ref{fig:comparison}, the biggest dilemma of previous RL-based agents is how to map an extremely long-horizon and complex goal to a sequence of lowest-level keyboard/mouse operations. To address this challenge, we propose our framework Ghost In the Minecraft (GITM)~\footnote{The name is chosen to pay tribute to the science fiction movie "Ghost in the Shell".}, which uses Large Language Model (LLM)-based agents as a new paradigm. Instead of direct mapping like RL agents, our LLM-based agents employ a hierarchical approach. It first breaks down the decompose goal into sub-goals, then into structured actions, and finally into keyboard/mouse  operations.
Such decomposition is similar to how humans solve complex problems in the real world, enabling mastery of Minecraft with efficiency orders of magnitude higher than that of RL. LLM can also leverage text-based knowledge and memory to quickly acquire the ability to interact with the environment and accomplish goals, offering immense learning efficiency improvements, unlimited scalability and representing a disruptive innovation compared with RL. Our GITM framework has the potential to revolutionize the path to generally capable agents.

Specifically, the proposed LLM-based agent consists of an LLM Decomposer, an LLM Planner, and an LLM Interface, which are responsible for the decomposition of sub-goals, structured actions, and keyboard/mouse operations, respectively. Given a goal in Minecraft, LLM Decomposer first decomposes it into a series of well-defined sub-goals according to the text-based knowledge collected from the Internet. Then, LLM Planner plans a sequence of structured actions for each sub-goal. The structured actions are defined with clear semantics and corresponding feedback, enabling LLMs to understand surrounding environments and make decisions at the cognitive level. LLM Planner also records and summarizes successful action lists into a text-based memory to enhance future planning. Finally, LLM Interface execute the structured actions to interact with the environment by processing raw keyboard/mouse input and receiving raw observations.

In this paper, we demonstrate the feasibility of employing Large Language Models (LLMs) to develop Generally Capable Agents (GCAs) within an open-world environment built from Minecraft. By exploiting the common sense and reasoning capabilities of LLMs for hierarchical goal decomposition, as well as utilizing text-based knowledge and memory, this paradigm shows the possibility of enabling agents to address a wide range of challenges within Minecraft and allowing them to effectively handle such open-world environment.
Consequently, our agent has surpassed all previous methods in achieving the \texttt{ObtainDiamond} goal (+47.5\% success rate).
Our agent also demonstrates superior learning efficiency compared to previous methods, reducing the number of environment interaction steps by more than 10,000$\times$.
Specifically, VPT~\cite{baker2022video} needs to be trained for 6,480 GPU days, DreamerV3~\cite{hafner2023mastering} needs to be trained for 17 GPU days, while our GITM does not require any GPUs and can be trained in just 2 days using a single CPU node with 32 CPU cores. More importantly, by obtaining all items in Minecraft Overworld as a milestone, this work represents a crucial first step towards achieving GCAs that can handle any task humans can accomplish in Minecraft.

%% file: srcs/03_relatedwork.tex
\vspace{-3mm}
\section{Related Work}
\vspace{-3mm}

\noindent\textbf{Minecraft agents} are intelligent programs that can perform various tasks within Minecraft world. 
Reinforcement learning has dominated this area for many years.
Some initial attempts have tried to use hierarchical RL~\cite{skrynnik2021hierarchical,mao2022seihai,lin2021juewu} or imitation learning~\cite{amiranashvili2020scaling} in MineRL competitions~\cite{milani2020retrospective,guss2021towards,kanervisto2022minerl}. 
Recently, with large-scale web data, VPT~\cite{baker2022video} builds a foundation model for Minecraft by learning from videos. 
Based on its success, many works~\cite{milani2023towards} have also explored to finetune foundation model with human feedback.
On the other hand, as Minecraft agents become increasingly proficient in handling simple tasks, the importance of multi-task learning becomes more prominent. 
Some previous works have adopted knowledge distillation~\cite{tessler2017deep} and curriculum learning~\cite{kanitscheider2021multi}, while recent works~\cite{fan2022minedojo,cai2023open} tried to construct a language-conditioned multi-task agent via feeding the goal description embedding into the model.

Recently, researchers have come to aware the extraordinary general planning ability for LLMs~\cite{huang2022language}.
Many works~\cite{huang2022inner,wang2023describe,yuan2023plan4mc} have leveraged LLMs for enhancing the high-level planning ability of minecraft agents.
Inner Monologue~\cite{huang2022inner} leveraged environment feedback to improve the planning ability of LLM. 
DEPS~\cite{wang2023describe} further extended this closed-loop interaction by introducing description, explainer and selector. 
Plan4MC~\cite{yuan2023plan4mc} pre-defined basic skills and instructed LLM to extract the relationship between skills to construct a skill graph.

Unlike previous RL-based or RL with LLM methods, our LLM-native approach brings the minecraft agent to another level both in efficiency and robustness by leveraging high-level action abstraction and text-based knowledge and memory.

\noindent\textbf{Large Language Models with Tools}
Extending the ability of LLMs by leveraging external tools have drawn a lot of attentions recently.
Several works~\cite{liang2022code,singh2022progprompt,driess2023palm} have explored to augment LLMs with robot perception and control abilities. 
Code as Polices\cite{liang2022code} tried to prompt LLM to generate codes that can drive robots. 
PaLM-E~\cite{driess2023palm} unified robot perception, instruction following, task planning and low-level control into a unified framework.
Another line of works tries to build external plugins around LLMs to enhance its ability.
Toolformer~\cite{schick2023toolformer} tries to teach LLMs to choose and use a wide range of tools like calculator and search engines and incorporate the results from tools into text generation.
HuggingGPT~\cite{shen2023hugginggpt} builds an agent for leveraging a combination of vision, language and audio models hosted on HuggingFace for completing user request.
API Bank~\cite{li2023api} proposes a syntheic benchmark suite for evaluating the how good LLMs are for using external tools.

Compared with these tool-augmented LLMs, our agents are tasks for much more complex goals in a high uncertain open-world.

%% file: srcs/04_method.tex
\vspace{-2mm}
\section{Method}
\vspace{-2mm}

Traditional RL-based agents struggle to develop generally capable agents in Minecraft. The core issue is that they attempt to map extremely long-horizon and complex goals directly to the lowest-level keyboard and mouse operations. To overcome this, we propose LLM-based agents in Fig.~\ref{fig:comparison} that utilize hierarchical goal decomposition. LLM Decomposer, LLM Planner, and LLM Interface are introduced to progressively decompose the task goal into sub-goals, structured actions, and keyboard/mouse operations. Moreover, LLM-based agents can leverage text-based knowledge and memory to quickly acquire the skills needed to master Minecraft. 

\begin{figure}[t]
 \small
 \centering
 \includegraphics[width=0.95\linewidth]{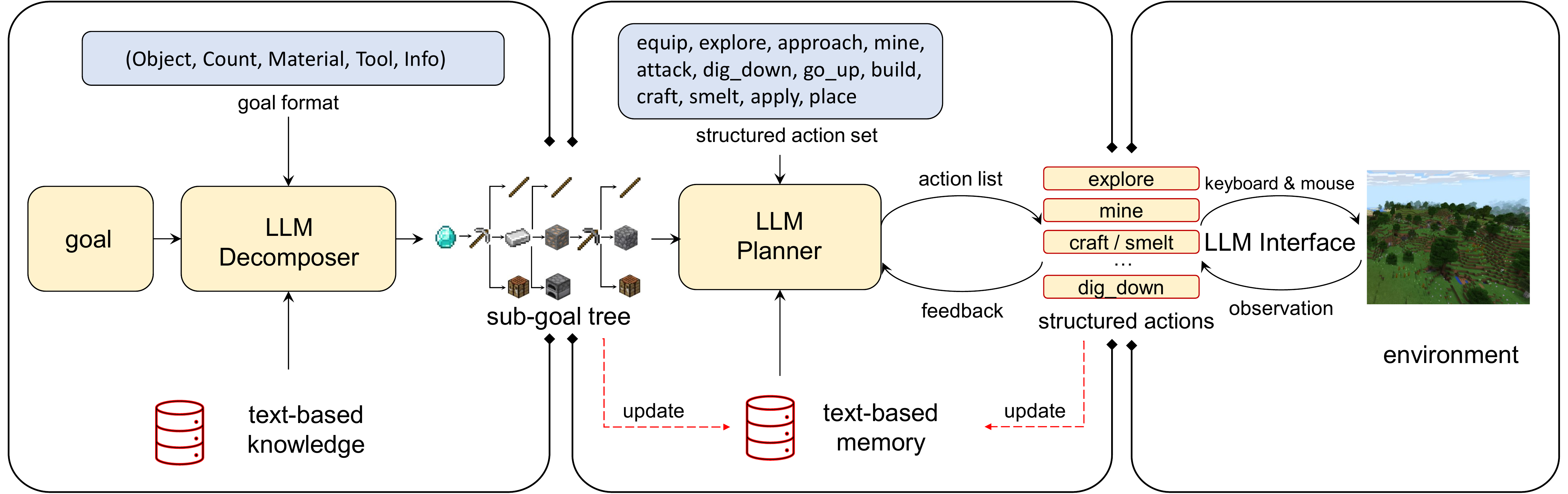}
 \caption{\textbf{Overview of our GITM.} Given a Minecraft goal, the LLM Decomposer divides the goal into a sub-goal tree. The LLM Planner then plans an action sequence for each sub-goal. Finally, the LLM Interface executes each action in the environment. Our LLM-based agents can be further enhanced by leveraging text-based knowledge and memory.}
\vspace{-3mm}
 \label{fig:overview}
\end{figure}

\vspace{-3mm}
\subsection{LLM Decomposer}

Rather than directly assigning the task goal to the agent and expecting a comprehensive and robust action plan, this work suggests the more practical strategy of decomposing the task goal into a series of more achievable sub-goals. By addressing each constituent sub-goal, the task goal can be progressively achieved.
To this end, an LLM Decomposer is proposed. Goals are fed to the decomposer and recursively decomposed into a sub-goal tree. Text-base knowledge provides the necessary information for decomposition.

\noindent\textbf{Goal Format.~}
Since we aim to obtain all items in Minecraft, all goals can be defined in the format of
\begin{equation}
\text{\texttt{(Object, Count, Material, Tool, Info)}},
\label{eq:goal}
\end{equation}
where ``Object'' denotes the target item, ``Count'' specifies the target quantity. ``Material'' and ``Tool'' refer to prerequisites needed to obtain the target item. ``Info'' stores the text-based knowledge related to this goal. Given a specific goal, its sentence embedding extracted from a pre-trained LLM is used to retrieve the most relevant text-based knowledge from an external knowledge base. Then, the LLM identifies the required material, tools, and related information from the gathered knowledge. The complete instructions for the LLM are described in Appendix.

\noindent\textbf{Recursive Decomposition. ~}
This goal format enables recursive decomposition of each goal into a sub-goal tree. Specifically, given a goal, all prerequisite items are listed as sub-goals, including materials, tools, and their corresponding quantities. Then, the recursive decomposition continues for each sub-goal until it has no prerequisites. After the decomposition is completed, the execution sequence of the sub-goals is planned through post-order traversal. Such goal decomposition significantly enhances the success rate of LLM planning, especially for goals necessitating long-horizon planning.

\noindent\textbf{Text-based Knowledge.~}
External knowledge is essential for automatic goal decomposition. We build an external knowledge base documented in text from the Minecraft Wiki on the Internet~\footnote{\url{https://minecraft-archive.fandom.com/wiki/Minecraft_Wiki}} and the item crafting/smelting recipes, providing an exhaustive source of knowledge about the Minecraft world. For instance, if we need to craft a wooden pickaxe, the item crafting recipe will indicate that the required materials are three planks and two sticks, and the necessary tool is a crafting table. It also provides information about the distribution of raw materials. For example, diamonds are frequently found in levels 10$\sim$12 underground.

\subsection{LLM Planner}

\begin{figure}[t]
  \small
  \centering
  \includegraphics[width=1.0\linewidth]{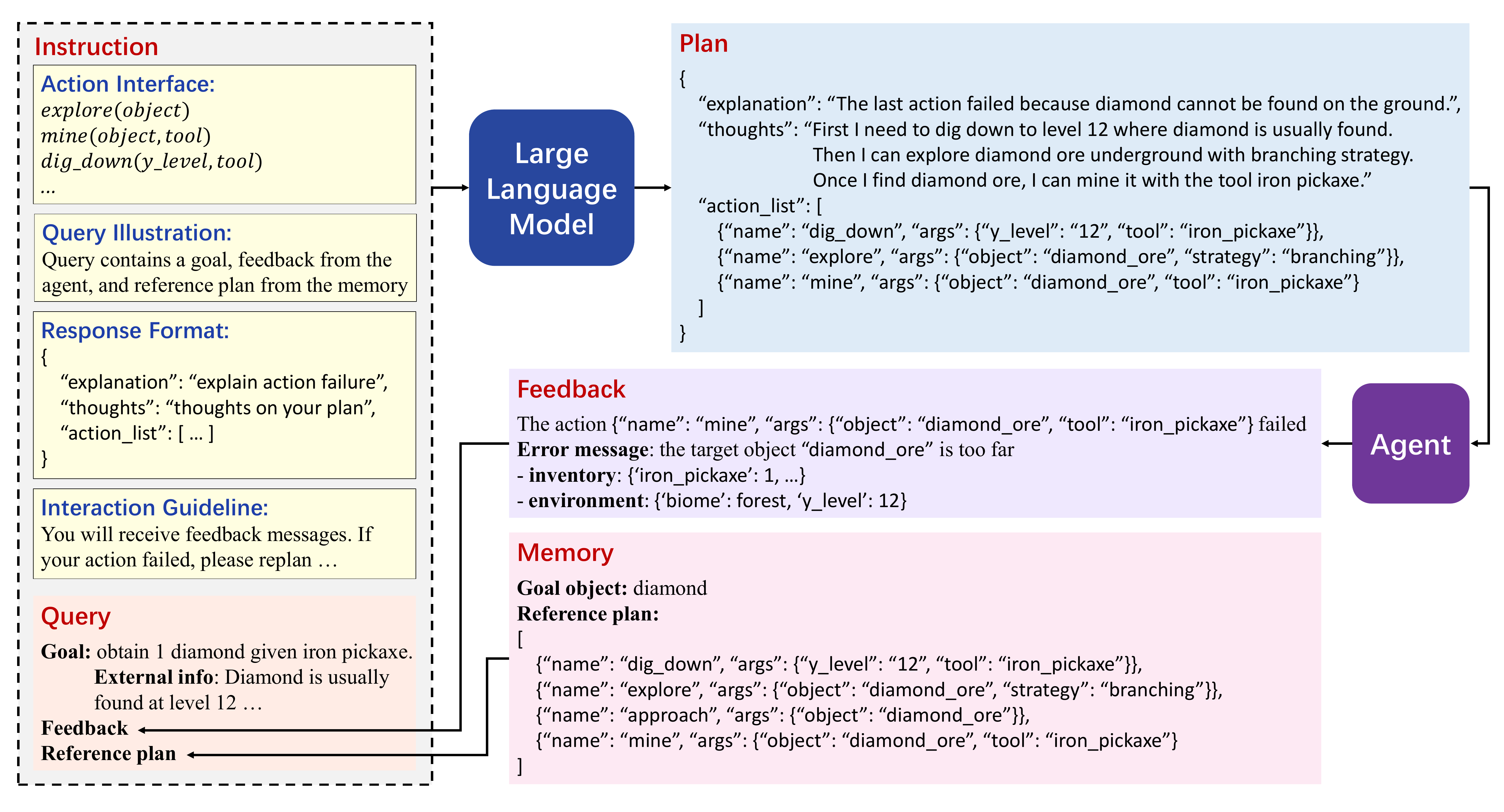}
  \caption{\textbf{Illustration of our planning process with the LLM Planner and the agent in the loop.} Given a specific goal, the planner generates plans with structured actions under the guidance of instruction, user query, previous feedback, and reference plan from memory. The agent executes the actions and provides feedback for the following planning. }
  \vspace{-2em}
  \label{fig:plan_process}
\end{figure}

LLMs excel at language understanding and reasoning but struggle with low-level control and multimodal perception. To leverage LLMs' strengths while addressing their limitations, we develop structured actions and feedback mechanisms as an abstract interface for them to manage agent-environment interaction.
We propose an LLM-based Planner to achieve goals in Minecraft. Given a goal, it generates structured actions to control agents, receives feedback, and revises plans accordingly. It also has a text memory that aids planning by providing solutions for frequent goals.

\noindent\textbf{Structured Actions.~}
The structured actions are designed with well-defined functions and clear semantics, enabling LLMs to make decisions at the cognitive level.
A structured action can be defined as follows:
\begin{equation}
\text{\texttt{(Name, Arguments, Description)}},
\label{eq:action}    
\end{equation}
The name and arguments defines the action we want the agent to execute, while the action description provides enough information for letting LLMs know when to choose the corresponding actions, as shown in Tab.~\ref{tab:method_atomic_actions}.

We extract the set of structured actions by leveraging the powerful reasoning capability of LLMs. Specifically, a pre-trained LLM is utilized to decompose the 3141 predefined tasks provided by MineDojo~\cite{fan2022minedojo} into action sequences. Instructions for guiding LLMs on action decomposition are provided in Appendix. 
Then, we extract the structured actions by selecting frequent actions and merging actions with similar functionalities. See Appendix for the set of structured actions. 

\input{tables/method-atomic_actions}

\noindent\textbf{Feedback Mechanism.~} Open-loop planning cannot guarantee success, especially in open-world environments, where agents might encounter unexpected events. Feedback is crucial to form an effective closed loop. Without appropriate feedback, the LLM has no information about the consequences of actions and may repeat failed action plans.
Feedback message is designed to present the agent's current state in the environment (\ie, inventory and environment), as well as the success and failure information for each executed actions, as shown in Fig.~\ref{fig:plan_process}. By incorporating this feedback message, the LLM can update its understanding of the environment, refine their strategies, and adapt their behavior accordingly.

\noindent\textbf{Planning.~}
Once the abstract interface is prepared, a pre-trained LLM is queried to generate goal-specific action sequence. This is achieved through carefully designed instructions and user queries, enabling the LLM to efficiently create and revise the plans. Fig.~\ref{fig:plan_process} illustrates the planning process. See Appendix for the full description.

\texttt{Instruction} specifies the guidelines that LLMs must follow when planning, including 1) \textit{Action Interface} provides functional descriptions of the structured actions and their parameters; 2) \textit{Query Illustration} clarifies the structure and meaning of user queries; 3) \textit{Response Format} requires LLM to return responses in the format of \{Explanation, Thought, Action List\}, where ``Explanation'' requires LLMs to explain the reason for action failure, ``Thought'' requires LLM to use natural language to plan before outputting action sequences as a chain-of-thought (CoT) mechanism~\cite{wei2022chain}, and ``Action List'' outputs a list of structured actions to be executed; 4) \textit{Interaction Guideline} guides LLMs to correct failed actions based on the feedback message, thus enabling the LLM to revise the plan.

\texttt{User Query} provides the specific query to LLMs for a given goal, including 1) \textit{Goal} represents the objective by text as ``Obtain \texttt{Count} \texttt{Item}, given \texttt{Material} and \texttt{Tool}. Extra info: \texttt{Info}'' according to Eq.~\eqref{eq:goal}; 2) \textit{Feedback} is the feedback information of the abstract interface; 3) \textit{Reference Plan} provides a common reference plan for the current goal retrieved from the text-base memory.

\noindent\textbf{Text-based Memory} is designed for LLM to maintain common reference plans for each encountered objective as experiential knowledge. LLMs acquire the experience about controlling agents and resolving specific situations through game play and agent interaction. Instead of starting from scratch every time, using prior experience allows LLMs to handle tasks more efficiently, a process similar to human skill improvement through practice.

To this end, we design a text-based memory mechanism for LLM to store and retrieve gained knowledge. 
Unlike the RL-based model, which stores knowledge in parameters, this textual knowledge is explicit, logical, and closely aligned with human thought processes. This allows for direct application to a wide range of similar tasks, leading to more efficient learning and improved generalization.

Specifically, during each game episode, once the goal is achieved, the entirely executed action list would be stored in memory. The LLM may achieve the same goal under various circumstances, resulting in a range of different plans. To identify a common reference plan suitable for general situations, essential actions from multiple plans are summarized. This summarization process is also implemented using LLMs (see Appendix for details). When encountering similar goals, the LLM creates new plans based on the summarized reference plans retrieved from memory. Successful action sequences from these new plans are also added to memory for future summarization. As the LLM-based Planner accumulates summaries, it becomes increasingly effective.

\vspace{-3mm}
\subsection{LLM Interface}
\vspace{-2mm}

Unlike the existing RL-based agents that directly control keyboard and mouse, LLM-based agents interact with the environment through structured actions and feedback messages. The LLM interface serves to implement structured actions as keyboard/mouse operations, and extract observations provided by the environment into feedback messages.

Structured actions can be implemented in various ways such as hand-written scripts or RL-learned models.
While RL-learned models have been employed in Minecraft previously, they were either broad in functionality but inefficient in practice, or too specific in functionality, limiting their applicability to general tasks and actions. 
Clarifying the capability boundary of RL-learned models is challenging.
Instead, in this work, we choose to implement structured actions using hand-written scripts. Since structured actions are well-defined and easy to implement, we can manually implement them based on observations (\eg, location, LiDAR, and voxel) and basic operations (\eg, move, jump, adjust camera angle, click left mouse button, and click right mouse button) provided by the MineDojo~\cite{fan2022minedojo} environment. 
See Appendix for details.

Feedback messages can be obtained directly from the environment. These include whether the structured action execution succeeded or failed. If the execution fails, the reason for the failure is additionally notified. It also includes the current state of the agent in the environment, including the items in the inventory, the current biome and depth, etc. See Appendix for details.

%% file: tables/method-atomic_actions.tex
\setlength{\tabcolsep}{4pt}
\begin{table}
    \centering
	\small
    \vspace{0.5em}
    \caption{\textbf{Examples of structured actions.} A structured action contains name and arguments for execution, as well as description to help LLMs understand and decide when to choose this action.}
    \vspace{0.5em}
    \resizebox{\textwidth}{!}{
    \begin{tabular}{l|l|l}
    \toprule
    Name & Arguments &  Description \\
    \midrule   
    \texttt{equip}& object & \makecell[l]{Equip the object from the inventory: used to equip equipment, including tools, weapons, and armor.} \\
     
    \texttt{explore}& object, strategy & \makecell[l]{Move around to find the object: used to find objects including block items and entities on the ground.} \\
 
    \texttt{approach}& object & \makecell[l]{Move close to a visible object: used to approach the object you want to attack or mine.} \\
 
    \texttt{mine/attack}& object, tool & \makecell[l]{Attack~/~Mine the object with the tool: used to attack~/~mine the object within reach.}  \\

    \texttt{dig\_down/go\_up}& ylevel, tool & \makecell[l]{Dig down~/~Go up with the tool: used to go down~/~up underground.} \\
 
    \texttt{build}& blueprint & \makecell[l]{Build according to a blueprint: used to place corresponding objects on locations according to a preset blueprint.} \\
 
    \texttt{craft/smelt}& object, tool, material & \makecell[l]{Craft~/~Smelt the object with the materials and tool: used to craft new object that is not in the inventory or is not enough.} \\
  
    \texttt{apply/place}& object, tool & \makecell[l]{Apply~/~Place the tool on the object: used to apply tools or place blocks.} \\
    \bottomrule
    \end{tabular}
    }
    \vspace{-0.5em}
    \label{tab:method_atomic_actions}
    \vspace{-1em}
\end{table}

%% file: srcs/05_experiments.tex
\section{Experiments}

\noindent\textbf{Task Definition and Metrics.} We measure the ability of GITM through item collection tasks. We only collect items could be found in the Overworld. We exclude items could only be obtained by trading with villagers, opening treasure chest or find a special structure on the map, using a tool enchanted with Silk Touch. This give us a total of 262 tasks.
For the assessment of our agent, we employ ``Coverage of the Overworld Technology Tree'' and``Success Rate'' as evaluation metrics.

\begin{figure}[t]
  \centering  
  \includegraphics[width=1\textwidth]{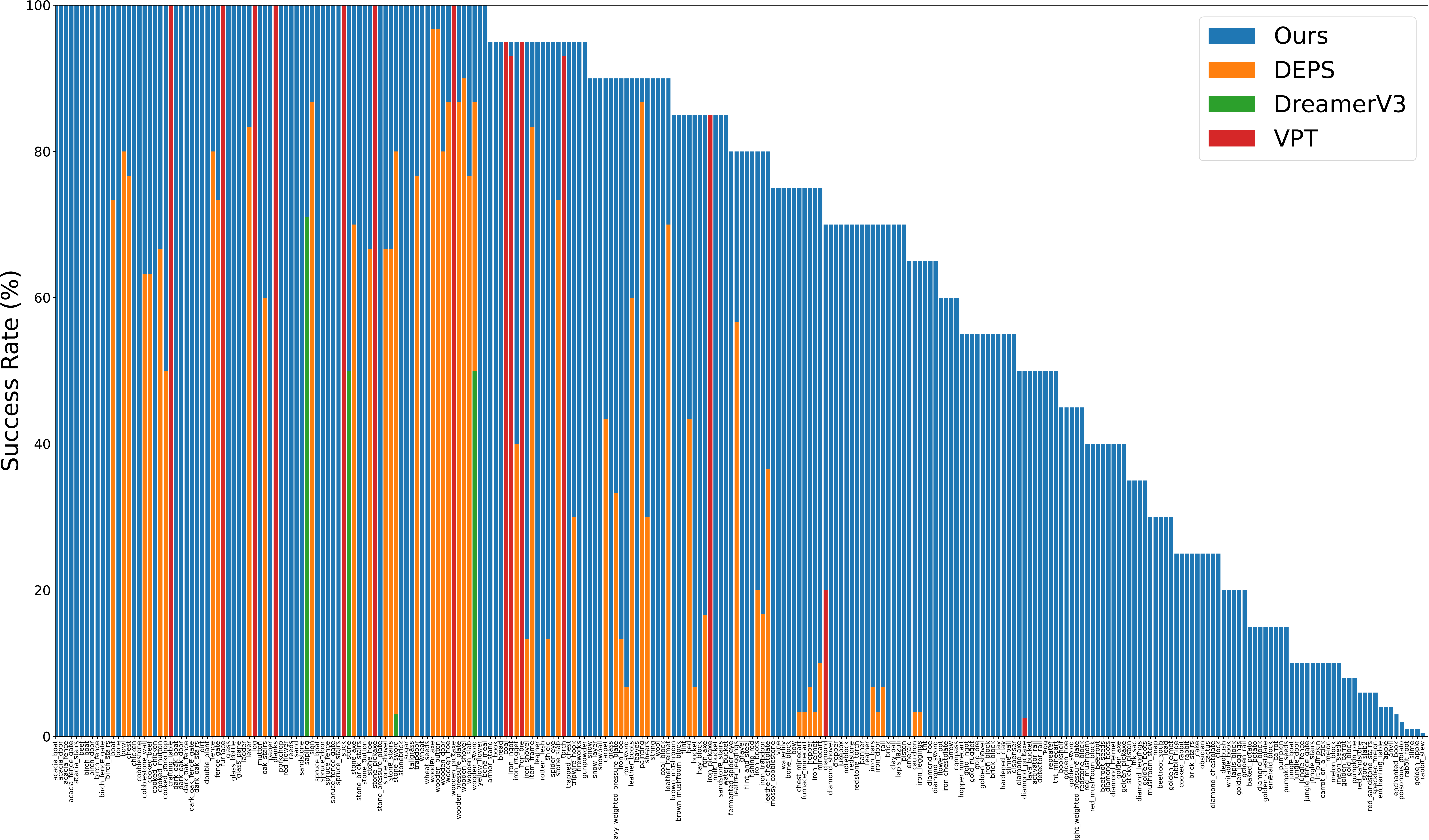}
  \caption{\textbf{Success rate for all items in the entire Minecraft Overworld Technology Tree}. The x axis lists all item names. We overlay the results from our GITM and the best results from baselines.}
  \label{fig:bar_chart}
\end{figure}

\subsection{Main Result}

\noindent\textbf{Unlocking the Entire Technology Tree by Obtaining All Items.~}
Compared with existing Minecraft agents~\cite{baker2022video,hafner2023mastering,wang2023describe} which mostly focuses on solving the \texttt{ObtainDiamond} task and could only unlock a limited part of the full technology tree (13/262 for Dreamerv3, 15/262 VPT, 69/262 for DEPS), our approach could collect all 262 items as shown in Fig.~\ref{fig:item_graph}. 
There are two major blockers for existing methods.
For RL-based methods like VPT~\cite{baker2022video} and DreamerV3~\cite{hafner2023mastering}, the goal item(diamond) is hard-coded into the model weights, which means there are no easy way to re-task the trained RL agents for collecting other items in the inference stage.
Moreover, the low training efficiency hinders them from solving extremely long-horizon tasks (e.g., obtaining a ``enchanted\_book''). 
For methods like DEPS~\cite{wang2023describe} that use an RL controller~\cite{cai2023open} and LLM planner still rely on pre-trained RL agents to execute specific subtasks (e.g. mining 1 ``cobblestone'') in the generated plan. So these approaches still suffer from the inability of RL-based methods alone to generalize to unseen tasks (e.g. obtaining ``lapis\_lazuli'').
In contrast, we extract a well-defined set of structured actions by using LLMs to decompose over 3000 predefined MineDojo tasks. This provides broad, open-world Minecraft capability. Combined with LLM planning, it enables solving more complex tasks than \texttt{ObtainDiamond} - which RL cannot achieve. Our knowledge bases also improve efficiency. To our knowledge, we present the first agent to unlock the entire Overworld technology tree - a level of open-world skill RL-based methods have not demonstrated.

\noindent\textbf{Success Rate for the Entire Technology Tree.~}
We show the success rate of our method for collecting all Overworld items in Fig.~\ref{fig:bar_chart}. 
Our methods could achieve 100\% success rate for simple tasks like collecting wooden tools.
It achieves non-zero success rates for all items which indicates a strong collecting capability.
The successful rate for collecting different items change smoothly for our agent, which showcase the robustness of our method against the highly uncertain open world environment.

\vspace{-3mm}
\subsection{Comparison with Other Minecraft Agents}

\begin{minipage}{\textwidth}
\begin{minipage}[c]{0.47\textwidth}
    \captionof{table}{\textbf{Comparison of our GITM with previous methods on \texttt{ObtainDiamond} challenge.}}
    \begin{tabular}{l|ccccc}
    \toprule
    \multirow{2}{*}{Method} & \multicolumn{5}{c}{Success Rate (\%)}\\
    & \mccraftingtable & \mcwoodenpickaxe & \mcstonepickaxe & \mcironpickaxe & \mcdiamond \\
    \midrule
    DreamerV3 & - & 50.0 & 3.0 & 0.01 & 0.01 \\
    DEPS & 90.0 & 80.0 & 73.3 & 10.0 & 0.6\\
    VPT & 100.0 & 100.0 & 100.0 & 85.0 & 20.0\\
    Our GITM & \textbf{100.0} & \textbf{100.0} & \textbf{100.0} & \textbf{95.0} & \textbf{67.5} \\
    \bottomrule
    \end{tabular}
    \label{tab:main_result}
\end{minipage}
\hfill
\begin{minipage}[c]{0.46\textwidth}
    \small
    \centering
    \includegraphics[width=0.8\linewidth]{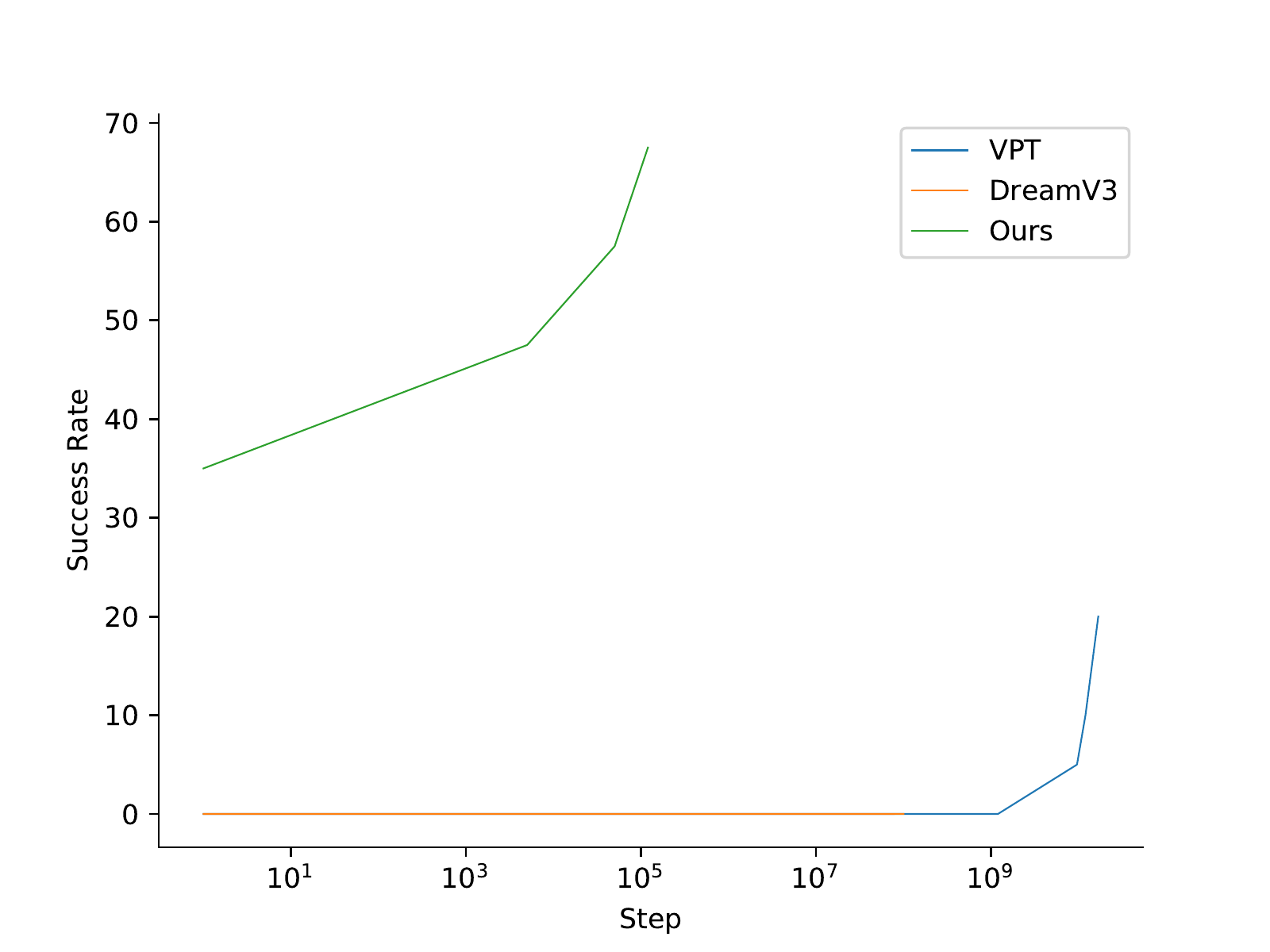}
    \vspace{-0.5em}
  \captionof{figure}{\hspace{0em}\textbf{Comparison of learning efficiency.}}
  \label{fig:learning_efficiency}
\end{minipage}
\end{minipage}

We compared our LLM-based method with three existing agents: VPT~\cite{baker2022video}, DreamerV3~\cite{hafner2023mastering}, and DEPS~\cite{wang2023describe} on the well known \texttt{ObtainDiamond} challenge, i.e, obtaining a diamond from scratch in Minecraft. 
Previous methods set different time limits of a single episode of game play (20 minutes for VPT, 30 minutes for Dreamerv3, and 10 minutes for DEPS). For fair comparison, we use the strictest limit of previous methods: 10 minutes (12,000 steps at 20Hz control).

\noindent\textbf{Success Rate for Obtaining Diamond and Other Items.~} Since VPT and Dreamerv3 are not targeted for collecting items other than diamond, we mainly compare our method with DEPS for items not related to obtain diamonds.
Overall, our GITM and VPT rank task difficulty similarly, but DEPS rankings severely fluctuate for tasks more complex than mining coal.
Dreamerv3 also behaves oddly by having an abnormally low success rate on tasks like obtaining a stone sword.
As shown in Fig.~\ref{tab:main_result}, most agents performs generally well for easy tasks relating to make wooden tools. VPT could even rival with our GITM for the success rate of obtaining iron axes. But for obtaining diamonds, our method wins over any other methods by 3.5 times on the succeess rate.

This giant improvement comes from the following two aspects:
First, we employ the strong long-term planning capability of LLMs to decompose the complex tasks into feasible sub-goals and tackle them step by step.
Second, our model can directly leverage external knowledge such as the suitable locations for mining ores, while RL models need to explore themselves and may not acquire reliable knowledge.

\noindent\textbf{Learning Efficiency.~} Besides measuring the success rate of each agents, we also compare the learning efficiency of our model with other learnable models.
Since DEPS uses a LLM-based planner without learning mechanism and a pre-trained RL-based controller, its performance could not improve with more episodes played and is excluded from the comparison here.

It usually takes tens of millions of steps to train an RL agent by updating parameters before its success rate starts to converges to meaningful non-zero numbers. 
However, the success rate for RL-based agents increases rather slowly even after them starts to converge.
On the contrary, the learning process of our LLM-based agent is considerably faster. As shown in Fig.~\ref{fig:learning_efficiency}, our method requires several orders less episodes than any other methods before doubling its initial success rate.
Moreover, our method is extremely sample efficient as our success rate raises from 35\% to 47.5\% by learning from the first five thousand steps. 
By just playing each task several times and summarize successful experience into the memory, the LLM-based agent can acquire explicit experiential knowledge and achieve significantly higher success rate.

\subsection{Ablation Study}
We conduct ablation experiments on the \texttt{ObtainDiamond} task. 
We set a time limit of 10 minutes of game play (12000 steps at the control frequency of 20Hz).
When leveraging goal decomposition, for each sub-goal, we set the maximum number of queries to LLM as 30, and exceeding the query limit will be judged as a failure.
For each setting, we run 40 games and calculate the success rate.
Tab.~\ref{tab:ablation_components} records the success rates of achieving the final goal diamond as well as the milestones in this goal, including crafting table, wooden pickaxe, stone pickaxe, and iron pickaxe.
\input{tables/ablation-component}

\noindent\textbf{Goal Decomposition.} 
Without goal decomposition, the planner can only accomplish several short-term tasks such as obtaining stone axes with rather low success rate of 5\%, which indicates the necessity of goal decomposition. Leveraging the powerful long-term planning capabilities of LLMs, the goals are decomposed into sub-goals feasible and practical for the planner, so the success rate for obtaining stone axes advances from 5\% to 67.5\% by leveraging goal decomposition alone.

\noindent\textbf{Feedback Message.}
Feedback contains the agent's state and the execution result of the actions, which helps the planner to understand and make another attempt to correct the mistakes in the previous and deal with special cases. This enables the planner to accomplish a broader range of goals with higher success rate. As shown in the 3rd row of Tab.~\ref{tab:ablation_components}, our agent gain the ability to collect diamond by combining feedback with goal decomposition.

\noindent\textbf{External Knowledge Base.}
External knowledge contains general rules, crafting recipes, and common tricks in Minecraft, such as the recipes for crafting iron ingot and iron pickaxe, the suitable location to find diamond ore, and the efficient way to get cobblestone. 
Providing the planner with this information greatly boosts the success rate of obtaining iron pickaxe and diamond, and the success rate of mining diamond increase by 7 times by learning from the knowledge base that diamonds are more likely to appear in specific levels.

\noindent\textbf{Text-based Memory.}
Leveraging the reference plan recorded in the memory, the planner can handle the task it has encountered more efficiently.
The success rates of obtaining iron pickaxe and diamond are 95.0\% and 67.5\%, surpassing the model without memory by 37.5\% and 32.5\%, respectively.

%% file: tables/ablation-component.tex
\setlength{\tabcolsep}{5pt}
\setlength{\doublerulesep}{2\arrayrulewidth}
\begin{table}[t]
    \centering
    \small
    \caption{\textbf{Ablation study.} The milestone items from left to right are crafting table \mccraftingtable, wooden pickaxe \mcwoodenpickaxe, stone pickaxe \mcstonepickaxe, iron pickaxe \mcironpickaxe, and diamond \mcdiamond. The success rate is calculated under time limit of 12000 steps (total) and query limit of 30 (each sub-goal). ``Goal Decomp.'' and ``External Info.'' indicates goal decomposition and external knowledge respectively.}
    \vspace{0.5em}
    \label{tab:ablation_components}
    \begin{tabular}{cccc|ccccc}
    \toprule
    \multirow{2}{*}{\makecell{Goal \\ Decomp.}} & \multirow{2}{*}{Feedback} & \multirow{2}{*}{\makecell{External \\ Info.}} & \multirow{2}{*}{Memory} & \multicolumn{5}{c}{Success Rate (\%)}\\
    & & & & \mccraftingtable & \mcwoodenpickaxe & \mcstonepickaxe & \mcironpickaxe & \mcdiamond \\
    \midrule
      & & & & 57.5 & 32.5 & 5.0 & 0.0 & 0.0 \\
     \checkmark & & & & 90.0 & 90.0 & 67.5 & 2.5 & 0.0 \\
     \checkmark & \checkmark & & & 97.5 & 95.0 & 77.5 & 20.0 & 5.0\\
     \checkmark & \checkmark & \checkmark & & 100.0 & 100.0 & 100.0 & 57.5 & 35.0 \\
     \checkmark & \checkmark & \checkmark & \checkmark & 100.0 & 100.0 & 100.0 & 95.0 & 67.5 \\
    \bottomrule
    \end{tabular}
\end{table}

%% file: srcs/06_conclusion.tex
\section{Conclusion}

We introduce the GITM framework, which utilizes Large Language Models (LLMs) for hierarchical decomposition of goals. GITM introduces LLM Decomposer, LLM Planner and LLM Interface to gradually decompose goals into sub-goals, structured actions and keyboard/mouse operations.
This work makes significant progress towards the \texttt{ObtainDiamond} goal, outperforming all previous methods by a significant margin (+47.5\% success rate). This proves the potential inefficiency and poorly scalability of Reinforcement Learning (RL) in Minecraft, breaking the traditional reliance on RL. Moreover, by obtaining all items in Minecraft Overworld, this research marks a critical step toward Generally Capable Agents (GCAs) that match human performance in Minecraft.

%% file: srcs/07_appendix.tex
\section{Implementation Details}
\subsection{LLM Decomposer}

We use \texttt{gpt-3.5-turbo} from OpenAI API~\footnote{\url{https://platform.openai.com/docs/api-reference}} for goal decomposition.
The prompt is shown as follows, which consists of two parts: instruction with the role of ``SYSTEM'' and query with the role of ``USER''.
The \texttt{\{object quantity\}}, \texttt{\{object name\}} and \texttt{\{related knowledge\}} are injectable slots that will be replace with corresponding texts before fed into the LLM.

\begin{tcolorbox}[breakable=false, boxrule={0.5pt}, sharp corners={all}]
\setlength{\parskip}{1ex}
\textbf{SYSTEM:}

You are an assistant for the game Minecraft.

I will give you some target object and some knowledge related to the object. Please write the obtaining of the object as a goal in the standard form.

The standard form of the goal is as follows:\\
\{\\
\hbox{\ \ \ \ }"object": "the name of the target object",\\
\hbox{\ \ \ \ }"count": "the target quantity",\\
\hbox{\ \ \ \ }"material": "the materials required for this goal, a dictionary in the form \{material\_name: material\_quantity\}. If no material is required, set it to None",\\
\hbox{\ \ \ \ }"tool": "the tool used for this goal. If multiple tools can be used for this goal, only write the most basic one. If no tool is required, set it to None",\\
\hbox{\ \ \ \ }"info": "the knowledge related to this goal"\\
\}

The information I will give you:\\
Target object: the name and the quantity of the target object\\
Knowledge: some knowledge related to the object.

Requirements:\\
1. You must generate the goal based on the provided knowledge instead of purely depending on your own knowledge.\\
2. The "info" should be as compact as possible, at most 3 sentences. The knowledge I give you may be raw texts from Wiki documents. Please extract and summarize important information instead of directly copying all the texts.

Goal Example:\\
\{
\hbox{\ \ \ \ }"object": "iron\_ore",\\
\hbox{\ \ \ \ }"count": 1,\\
\hbox{\ \ \ \ }"material": None,\\
\hbox{\ \ \ \ }"tool": "stone\_pickaxe",\\
\hbox{\ \ \ \ }"info": "iron ore is obtained by mining iron ore. iron ore is most found in level 53. iron ore can only be mined with a stone pickaxe or better; using a wooden or gold pickaxe will yield nothing."\\
\}\\
\{\\
\hbox{\ \ \ \ }"object": "wooden\_pickaxe",\\
\hbox{\ \ \ \ }"count": 1,\\
\hbox{\ \ \ \ }"material": \{"planks": 3, "stick": 2\},\\
\hbox{\ \ \ \ }"tool": "crafting\_table",\\
\hbox{\ \ \ \ }"info": "wooden pickaxe can be crafted with 3 planks and 2 stick as the material and crafting table as the tool."\\
\}\newline
\newline
\textbf{USER:}

Target object: \texttt{\{object quantity\}} \texttt{\{object name\}}\\
Knowledge: \texttt{\{related knowledge\}}
\end{tcolorbox}

The recursive decomposition generates a sub-goal tree starting from the final goal object as the root node:
if a goal has some prerequisites (materials or tools), for each required material or tool, we add a child node representing the goal of obtaining that material or tool, and then recursively decompose the child node, until there is no more prerequisites.
The related knowledge is from:
1) Crafting/smelting recipes in MineDojo~\cite{fan2022minedojo}, written in the form ``Crafting \texttt{\{quantity\}} \texttt{\{object\}} requires \texttt{\{material\}} as the material and \texttt{\{tool\}} as the tool''; 
2) Wiki on the Internet~\footnote{\url{https://minecraft-archive.fandom.com/wiki/Minecraft_Wiki}}. We extract the paragraphs with keywords ``obtaining'', ``mining'', ``sources'', etc.

\subsection{LLM Interface}
\noindent\textbf{Instruction for Extracting Structured Actions.}
To extract structured actions, we first ask LLM to generate a tree-structured action planning for each of the 3141 predefined tasks provided by
MineDojo, and then converts each action step into a \texttt{(verb, object, tool, material)} tuple.
During decomposition, it is essential to ensure actions are neither too broad nor too specific.
We adjusted the depth of the action decomposition tree to achieve balance, and empirically set the depth as 2 to meet our requirements.

Specifically, we use \texttt{gpt-3.5-turbo} from OpenAI API to generate the structured actions. We add the following instruction to the content of ``SYSTEM'' role to generate the tree-structured plan. 
We add the goal description, \eg, "find material and craft a iron pickaxe", to the content of ``USER'' role and then asks LLM to response according to the requirements.

\begin{tcolorbox}[breakable=false, boxrule={0.5pt}, sharp corners={all}]
\setlength{\parskip}{1ex}
\textbf{SYSTEM:}\\
You serve as an assistant that helps me play Minecraft.

I will give you my goal in the game, please break it down as a tree-structure plan to achieve this goal. 

The requirements of the tree-structure plan are:

1. The plan tree should be exactly of depth 2.\\
2. Describe each step in one line.\\
3. You should index the two levels like '1.', '1.1.', '1.2.', '2.', '2.1.', etc.\\
4. The sub-goals at the bottom level should be basic actions so that I can easily execute them in the game.
\newline
\newline
\textbf{USER:}\\
The goal is to \{\texttt{goal description}\}. Generate the plan according to the requirements.
\end{tcolorbox}

After that, we extract the action tuple from each sentence of the leaf nodes. We use the following instruction as the content of ``SYSTEM'' role to extract the tuple, and add the sentence to the content of ``USER'' role.

\begin{tcolorbox}[breakable=false, boxrule={0.5pt}, sharp corners={all}]
\setlength{\parskip}{1ex}
\textbf{SYSTEM:}\\
You serve as an assistant that helps me play Minecraft.

I will give you a sentence. Please convert this sentence into one or several actions according to the following instructions.

Each action should be a tuple of four items, written in the form ('verb', 'object', 'tools', 'materials')

'verb' is the verb of this action.\\
'object' refers to the target object of the action.\\
'tools' specifies the tools required for the action.\\
'material' specifies the materials required for the action.\\
If some of the items are not required, set them to be 'None'.
\newline
\newline
\textbf{USER:}\\
The sentence is \{\texttt{sentence}\}. Generate the action tuple according to the requirements.
\end{tcolorbox}

Then, we extract the structured actions by selecting frequent actions and merging actions with similar functionalities. The set of structured actions is \texttt{\{equip, explore, approach, mine/attack, dig\_down, go\_up, build, craft/smelt, apply\}}. Note that we disregard more detailed action decomposition for \texttt{attack} and \texttt{build} to remove overly detailed short-term actions and focus on long-term task completion. 

\noindent\textbf{Action Implementation.}
The observation of the action contains LiDAR rays with an interval of 5 degrees in the horizon and vertical direction for locating objects, and voxels with 10 units radius only for navigation, inventory, life status, and agent location status (X-ray cheating is carefully avoided). RGB is not used in our implementation, although it provides more information than LiDAR rays. For example, the biome, and category of the dropping item can not be identified by LiDAR rays. Some objects may also be missed by LiDAR due to sparseness of LiDAR rays. 
We also set the breaking speed to 100 and strength to 100, mainly following \cite{hafner2023mastering}.
The detailed implementation of each structured action is as follows:
\begin{itemize}[leftmargin=1em]
\item \texttt{equip}: The equip action calls the environment API to equip the required object. The action succeeds when the API returns success. The action fails when the object is not in inventory or the equip API returns failure. 

\item \texttt{explore}: The explore action traverses the world until object is visible. This action regards the world as a chessboard, and each node on the chessboard is the center point of a 20$\times$20 units area.
Two strategies are implemented depending on whether the agent is on the ground or not. When the agent is on the ground, the BFS explore will be adopted. When the agent is under the ground, mainly for exploring ore, the DFS explore will be adopted. In the DFS exploration, the agent will break the blocks to form a mine road with width of 1 and height of 2. The action succeeds when the object is visible. The action fails when the explore exceeds a preset steps of 10,000 but the required object is not found.

\item \texttt{approach}: The approach action finds the nearest visible required object and walks towards the object. We adopt $A^*$ algorithm for finding path. 
The $A^*$ algorithm can jump, translate and fall in four directions of north, south, east and west. We also allow the agent to jump while placing a block under the agent for ascent.
If the object is out of the voxel observation range, $A^*$ algorithm is iteratively applied to find the location nearest to the object. The action succeeds when the $\ell^{\infty}$ norm distance between the object and agent is less than 2. 
The action fails when there is no required object visible or no path can be found to walk close to the object.

\item \texttt{mine/attack}: The mine/attack action uses the keyboard attack API with the tools to attack the object. Only visible object could be mined or attacked. The object of mine should be blocks, and the agent will continue mining the block until it is broken. The object of attack should be entities, and the agent will iteratively approach and attack the entity until it is killed. After the block is broken or the entity is killed, if there are items dropped by them, the agent will approach the items to collect them. The action succeeds when the block is broken or the entity is killed. The action fails when there is no visible object, no required tools is in inventory, or the visible object is out of attack range.

\item \texttt{dig\_down}: The dig\_down action iteratively breaks the block underfoot with the tool until the required ylevel is reached. If the agent is on the ground, before digging down, current location is stored for going up action. After the action succeeds, the state of the agent is set to under ground. The action succeeds when the required ylevel is reached. The action fails when it exceeds the reset max steps 10,000 or no required tool is in inventory.   

\item \texttt{go\_up}: The agent will first go back to the location stored by dig\_down. Then, the go\_up action puts dirt blocks underfoot to raise the agent. After the action is finished, the state of agent is set to on the ground. The action succeeds when the pre-stored location is reached. The action fails when the walk fails, exceeds the reset max steps 10,000 or there is no required tool in inventory.

\item \texttt{build}: The build action places the required blocks according to a given blueprint from bottom to up. The action succeeds when all blocks have been placed. The action fails when there are no enough materials in inventory or it is invalid to place some blocks.

\item \texttt{craft/smelt}: The action calls the environment API to craft/smelt the required object. The action succeeds when the required object is obtained. The actions fails when there are no enough materials in inventory or the agent is unable to place the crafting table/furnace or the API fails.

\item \texttt{apply}: The apply action calls the keyboard use API, and applies the specific tool to the object, \eg, applying the bucket on water to obtain water bucket. The action succeeds when the API returns success. The action fails when there is no visible object, no tool in inventory or the API fails.
\end{itemize}

\noindent\textbf{Feedback Message.}
After the execution of each action, we will get feedback from the structured actions. The feedback will refresh the agent's state in Sec.~\ref{subsubsec:query}, including current inventory, biome, ylevel and on/under the ground status. The feedback will also contain the success/fail message from these action, as well as the inventory change during the action.

\subsection{LLM Planner}
Here we present the prompt for planning with LLM.
We also use \texttt{gpt-3.5-turbo} from OpenAI API as the LLM planner.
The model accepts inputs in form of a chat, i.e., the prompt is a dialogue consisting of several messages, each of which contains a role and the content.
We set the \texttt{Instruction} with the role ``SYSTEM'' at the beginning, and use the \texttt{User Query} with the role ``USER'' to query the LLM for response.
The content of the \texttt{Instruction} and \texttt{User Query} are as follows.

\subsubsection{Instruction}
\label{sec:instruction_prompt}
\begin{tcolorbox}[breakable=true, boxrule={0.5pt}, sharp corners={all}]
\setlength{\parskip}{1ex}
\textbf{SYSTEM:}

You serve as an assistant that helps me play the game Minecraft.

I will give you a goal in the game. Please think of a plan to achieve the goal, and then write a sequence of actions to realize the plan. The requirements and instructions are as follows:

1. You can only use the following functions. Don't make plans purely based on your experience, think about how to use these functions.

\texttt{explore(object, strategy)}\\
Move around to find the object with the strategy: used to find objects including block items and entities. This action is finished once the object is visible (maybe at the distance).\\
Augments:\\
- object: a string, the object to explore.\\
- strategy: a string, the strategy for exploration.

\texttt{approach(object)}\\
Move close to a visible object: used to approach the object you want to attack or mine. It may fail if the target object is not accessible.\\
Augments:\\
- object: a string, the object to approach.

\texttt{craft(object, materials, tool)}\\
Craft the object with the materials and tool: used for crafting new object that is not in the inventory or is not enough. The required materials must be in the inventory and will be consumed, and the newly crafted objects will be added to the inventory. The tools like the crafting table and furnace should be in the inventory and this action will directly use them. Don't try to place or approach the crafting table or furnace, you will get failed since this action does not support using tools placed on the ground. You don't need to collect the items after crafting. If the quantity you require is more than a unit, this action will craft the objects one unit by one unit. If the materials run out halfway through, this action will stop, and you will only get part of the objects you want that have been crafted.\\
Augments:\\
- object: a dict, whose key is the name of the object and value is the object quantity.\\
- materials: a dict, whose keys are the names of the materials and values are the quantities.\\
- tool: a string, the tool used for crafting. Set to null if no tool is required.

\texttt{mine(object, tool)}\\
Mine the object with the tool: can only mine the object within reach, cannot mine object from a distance. If there are enough objects within reach, this action will mine as many as you specify. The obtained objects will be added to the inventory.\\
Augments:\\
- object: a string, the object to mine.\\
- tool: a string, the tool used for mining. Set to null if no tool is required.

\texttt{attack(object, tool)}\\
Attack the object with the tool: used to attack the object within reach. This action will keep track of and attack the object until it is killed.\\
Augments:\\
- object: a string, the object to attack.\\
- tool: a string, the tool used for mining. Set to null if no tool is required.

\texttt{equip(object)}\\
Equip the object from the inventory: used to equip equipment, including tools, weapons, and armor. The object must be in the inventory and belong to the items for equipping.\\
Augments:\\
- object: a string, the object to equip.

\texttt{digdown(object, tool)}\\
Dig down to the y-level with the tool: the only action you can take if you want to go underground for mining some ore.\\
Augments:\\
- object: an int, the y-level (absolute y coordinate) to dig to.\\
- tool: a string, the tool used for digging. Set to null if no tool is required.

\texttt{go\_back\_to\_ground(tool)}\\
Go back to the ground from underground: the only action you can take for going back to the ground if you are underground.\\
Augments:\\
- tool: a string, the tool used for digging. Set to null if no tool is required.

\texttt{apply(object, tool)}\\
Apply the tool on the object: used for fetching water, milk, lava with the tool bucket, pooling water or lava to the object with the tool water bucket or lava bucket, shearing sheep with the tool shears, blocking attacks with the tool shield.\\
Augments:\\
- object: a string, the object to apply to.\\
- tool: a string, the tool used to apply.

2. You cannot define any new function. Note that the "Generated structures" world creation option is turned off.

3. There is an inventory that stores all the objects I have. It is not an entity, but objects can be added to it or retrieved from it anytime at anywhere without specific actions. The mined or crafted objects will be added to this inventory, and the materials and tools to use are also from this inventory. Objects in the inventory can be directly used. Don't write the code to obtain them. If you plan to use some object not in the inventory, you should first plan to obtain it. You can view the inventory as one of my states, and it is written in form of a dictionary whose keys are the name of the objects I have and the values are their quantities.

4. You will get the following information about my current state:\\
- inventory: a dict representing the inventory mentioned above, whose keys are the name of the objects and the values are their quantities\\
- environment: a string including my surrounding biome, the y-level of my current location, and whether I am on the ground or underground\\
Pay attention to this information. Choose the easiest way to achieve the goal conditioned on my current state. Do not provide options, always make the final decision.

5. You must describe your thoughts on the plan in natural language at the beginning. After that, you should write all the actions together. The response should follow the format:\\
\{\\
\hbox{\ \ \ \ }"explanation": "explain why the last action failed, set to null for the first planning",\\
\hbox{\ \ \ \ }"thoughts": "Your thoughts on the plan in natural languag",\\
\hbox{\ \ \ \ }"action\_list": [\\
\hbox{\ \ \ \ \ \ \ \ }\{"name": "action name", "args": \{"arg name": value\}, "expectation": "describe the expected results of this action"\},\\
\hbox{\ \ \ \ \ \ \ \ }\{"name": "action name", "args": \{"arg name": value\}, "expectation": "describe the expected results of this action"\},\\
\hbox{\ \ \ \ \ \ \ \ }\{"name": "action name", "args": \{"arg name": value\}, "expectation": "describe the expected results of this action"\}\\
\hbox{\ \ \ \ }]\\
\}\\
The action\_list can contain arbitrary number of actions. The args of each action should correspond to the type mentioned in the Arguments part.
Remember to add ```{dict}``` at the beginning and the end of the dict.
Ensure that you response can be parsed by Python json.loads

6. I will execute your code step by step and give you feedback. If some action fails, I will stop at that action and will not execute its following actions. The feedback will include error messages about the failed action. At that time, you should replan and write the new code just starting from that failed action.
\end{tcolorbox}

\subsubsection{User Query}
\label{subsubsec:query}
\begin{tcolorbox}[breakable=true, boxrule={0.5pt}, sharp corners={all}]
\setlength{\parskip}{1ex}
\textbf{USER:}

My current state:\\
- inventory: \texttt{\{inventory\}}\\
- environment: \texttt{\{environment\}}

The goal is to \texttt{\{goal\}}.

Here is one plan to achieve similar goal for reference: \texttt{\{reference plan\}}.

Begin your plan. Remember to follow the response format.\\
\textit{or} Action \texttt{\{successful action\}} succeeded, and \texttt{\{feedback message\}}. Continue your plan. Do not repeat successful action. Remember to follow the response format.\\
\textit{or} Action \texttt{\{failed action\}} failed, because \texttt{\{feedback message\}}. Revise your plan from the failed action. Remember to follow the response format.
\end{tcolorbox}

\subsection{Memory}
\subsubsection{Learning Process}
We maintain the text-based memory with a dictionary, whose keys are sub-goals and values are lists of successful action sequences for the corresponding sub-goals. The construction and update of the memory are through the following learning process:
\begin{itemize}[leftmargin=1em]
\item When encountering a new sub-goal that is not in the memory, the LLM planner creates plans without reference. Once the sub-goal is achieved, the entirely executed action sequence would be stored into the memory.
\item When encountering a sub-goal with memory, the first action sequence in the recording list for this goal is retrieved as the reference plan, with which the LLM planner tries to achieve the goal.
If it succeeds, the new executed action sequence will be added to the last of the recording list.
\item For each sub-goal, once the number of action sequences recorded in its list reaches $N$, we pop all the $N$ sequences and use LLM to summarize them into a common plan solution suitable for various scenarios, which is then put first in the list.
$N$ is set to 5 in all our experiments.
\end{itemize}

To learn the memory for obtaining all items, starting from scratch each time would take a long time. 
In addition, it is necessary to avoid spending the most of time on learning simple tasks and not investing enough in learning difficult tasks.
To improve the learning efficiency, we suggest to study the sub-goals individually one by one.
We first use our LLM Decomposer to generate sub-goal trees for all items, acquiring the set of all sub-goals involved.
Then for each sub-goal, the LLM planner plays multiple times given its prerequisites including the required materials and tools.
The learning process of the sub-goal is finished once we obtain $N=5$ successful action sequences and summarize them into one common plan solution for reference.

\subsubsection{Implementation of Memory Summarization}
We also use \texttt{gpt-3.5-turbo} from OpenAI API for memory summarization but in a different dialogue. 
We use the following prompt to instruct the summarization with the role ``SYSTEM''.
The slot \texttt{\{action description\}} is replaced with the same descriptions of interfaces of the structured actions as Sec. \ref{sec:instruction_prompt}.
We list all the action sequences to be summarized in the query with the role ``USER'', which is fed into the LLM for response.

\begin{tcolorbox}[boxrule={0.5pt}, sharp corners={all}]
\setlength{\parskip}{1ex}
\textbf{SYSTEM:}

You serve as an assistant that helps me play the game Minecraft.

I am using a set of actions to achieve goals in the game Minecraft. I have recorded several action sequences successfully achieving a goal in a certain state. I will give you the goal, the state, and the sequences later. Please summarize the multiple action sequences into a single action sequence as a universal reference to achieve the goal given that certain state. Here are the instructions:

1. Each action sequence is a sequence of the following actions:

\texttt{\{action description\}}

2. The action sequences before and after summarization are always conditioned on the given state, i.e., the actions are taken in that certain state to achieve the goal. I will describe the state in the following form:
State:
- inventory: a dict whose keys are the name of the objects and the values are their quantities. This inventory stores all the objects I have.
- environment: a dict including my surrounding biome and whether I am on the ground or underground.

3. The action sequence you summarize should be able to achieve the goal in general cases without specific modification. Every necessary action should be included, even though it does not appear in some sequences because I manually skipped it in some lucky cases. The actions redundant or irrelevant to the goal should be filtered out. The corner cases, such as success by luck and dealing with contingencies, should not be summarized into the final sequence.

4. You should describe your thoughts on summarization in natural language at the beginning. After that, give me the summarized action sequence as a list in JSON format. Your response should follow this form:

Thoughts: "Your thoughts and descriptions of your summarization"\\
Summarized action sequence:\newline
[\newline
\hbox{\ \ \ \ }\{"name": "action name", "args": \{"arg name": value\}, "expectation": "describe the expected results of this action"\},\\
\hbox{\ \ \ \ }\{"name": "action name", "args": \{"arg name": value\}, "expectation": "describe the expected results of this action"\},\\
\hbox{\ \ \ \ }\{"name": "action name", "args": \{"arg name": value\}, "expectation": "describe the expected results of this action"\}\\
]
\end{tcolorbox}

\section{Results of All Items}
We provide the success rate of all items in the entire Minecraft Overworld Technology Tree in Tab.~\ref{tab:appendix-all-item}. We have attached a video of obtaining a \textit{diamond} in the supplementary materials.

\noindent\textbf{Experiment Setting.}
Considering the large number of items, including those difficult to be obtained, we implemented an incremental testing strategy. This strategy is designed to keep the testing costs within a reasonable range, while also accounting for the rarity of certain items. We avoided a uniform increase in the number of tests across all items to accommodate the hardest-to-obtain ones, which would have resulted in prohibitive testing costs. Instead, we employed a incremental testing process.

For each item, we begin with 20 games. If the success count is less than or equal to 1, we increase to 50 games. If the success count remains less than or equal to 1, we further increase to 100, and eventually 200 games. This testing continues until the success count finally exceeds 1, or we complete 200 games.
By following this efficient strategy, we ensure a cost-effective and reliable evaluation of each item, regardless of its availability.
Moreover, because some items need long-term planning and crafting chain, we do not set restrictions on the time limit or query limit.


\noindent\textbf{Exploring Biome.} Biomes can be a key factor that strongly influences the success rate. Some items, like cactus, pumpkin, or melon, can only be found in specific biomes. The distribution of biomes highly limits the success rate of some items.

\input{tables/appendix-all-item-success-rate}

\section{Supplementary Ablations} 

\input{tables/appendix-ablation}

We make a more detailed comparison between our GITM with RL-based methods in Tab.~\ref{tab:appendix_ablation}. The most straightforward pipeline is to directly map the goal into keyboard/mouse operations. We gradually add goal decomposition and structured action stages into the pipeline, and ablate the use of RL-based models or LLM.

\noindent\textbf{Implementation Details.} We can only find open-sourced RL models from VPT~\cite{baker2022video} and DEPS~\cite{wang2023describe}, so they are adopted for the ablation. VPT model is specifically trained for the \texttt{ObtainDiamond} challenge, while DEPS model can use goal description as input to guide the model's output. We refer to them as specialist RL model and goal-conditioned RL model, respectively.
As for the use of LLM Planner, we note that if structured action is not used, LLM Planner will be inevitably asked to output reasonable keyboard/mouse operations. However, LLM Planner does not have access to environment observations, so it cannot directly output reasonable keyboard/mouse operations.

\noindent\textbf{Direct Mapping.} See Tab.~\ref{tab:appendix_ablation}(a)(b)(c). It is hard to directly mapping the long-horizon goal into reasonable keyboard/mouse operations. While a specialist RL model (\ie, VPT) can deliver promising results, it requires large amount of data and computational resources to train such a model~\cite{baker2022video} (720 V100 GPUs for 9 days). Moreover, a different goal will require further training of the specialist RL model, limiting the versatility of this paradigm.
The goal conditional RL model (\ie, DEPS) cannot achieve the goal, because the model~\cite{wang2023describe} we have access to is not generalizable to all scenarios. If only the final goal is given, it will ignore preconditions, such as not crafting the necessary iron pickaxe when mining diamonds.
LLM also fails to accomplish the goal. The primary reason is that it can not handle environment observation and keyboard/mouse operations well.

\noindent\textbf{Structured Action.} We design structured actions to interact with the environment, and provide an abstract interface. Tab.~\ref{tab:appendix_ablation}(f) shows that adding structured action significantly improves LLM's performance. This is because structured actions can deal with environment observations and keyboard/mouse operations more precisely, unleashing the reasoning potential of LLM.
We are not aware of a RL model using structured actions currently. It is possible for structure actions to enhance the RL model as well, and we will explore it in the future work.

\noindent\textbf{Goal Decomposition.} Decomposing the goal into sub-goals can simplify the whole task. Tab.~\ref{tab:appendix_ablation}(b)(d) and Tab.~\ref{tab:appendix_ablation}(f)(g) show its effectiveness for both goal-conditioned RL model and our method. By exploiting goal decomposition, it is possible for our method to accomplish long-term tasks with high success rate.

\noindent\textbf{Comparison between RL-based methods.} We also note the paradigm shift from traditional RL-based methods to our GITM leads to a great performance boost. Comparing Tab.~\ref{tab:appendix_ablation}(d)(g), where we only change the goal-conditioned RL model to LLM with strutured actions, our method significantly outperforms the RL model.

\section{\texttt{ObtainDiamond}}

\begin{figure}
    \centering
    \includegraphics[width=0.9\textwidth]{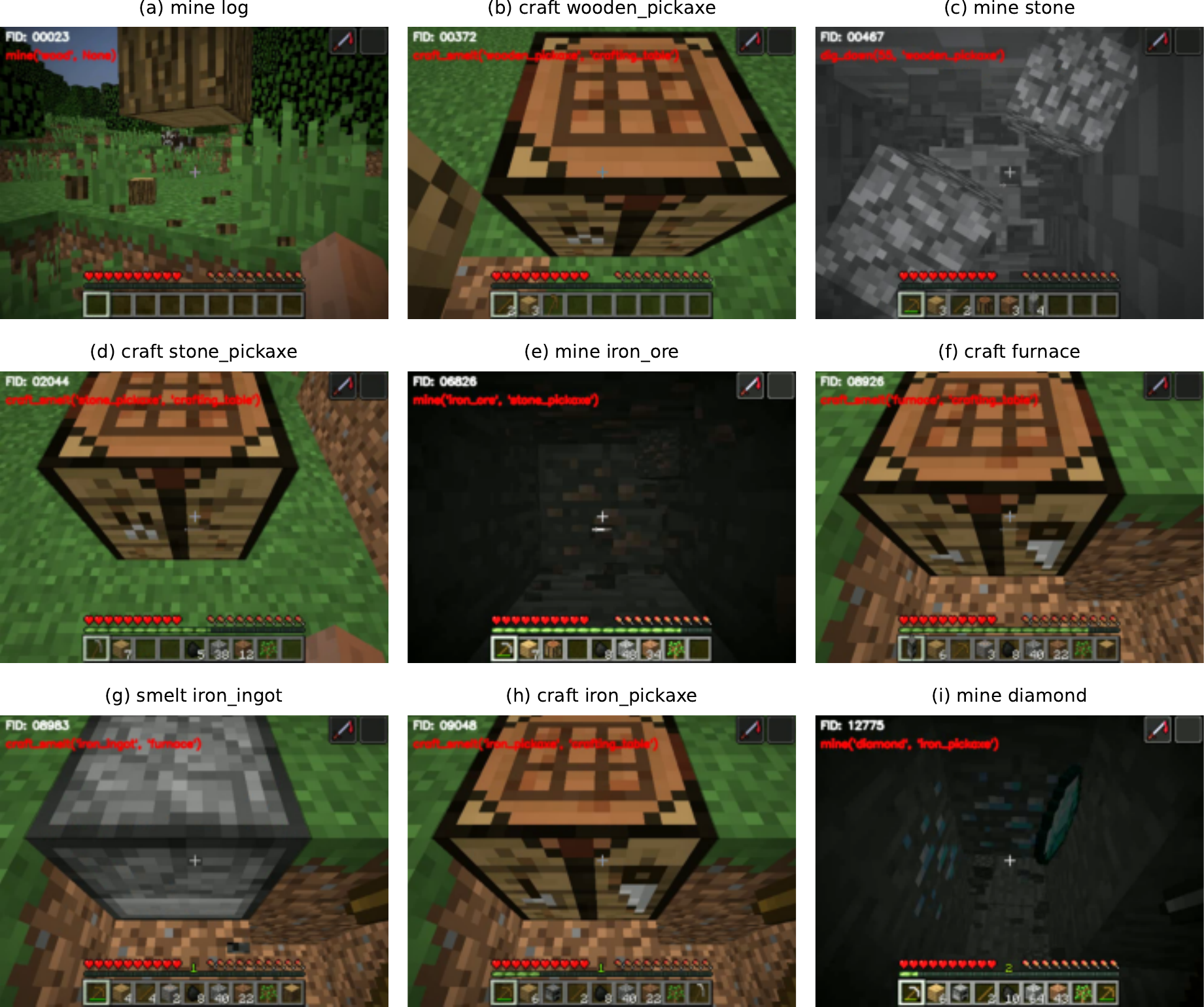}
    \caption{\textbf{A case of the popular \texttt{ObtainDiamond} challenge.} Figure(e)(i) are enhanced in brightness for better display.}
    \label{fig:appendix_case_study}
    \vspace{-1em}
\end{figure}

We demonstrate a case of the popular \texttt{ObtainDiamond} challenge in Fig.~\ref{fig:appendix_case_study}. During the process, the agent have to collect materials, \ie, log, stone and iron ore, as shown in Fig~\ref{fig:appendix_case_study}(a)(c)(e). Necessary tools, \ie, wooden pickaxe, stone pickaxe, furnace and iron pickaxe are also crafted in Fig~\ref{fig:appendix_case_study}(b)(d)(f)(h). Finally the diamond is obtained in Fig~\ref{fig:appendix_case_study}(i).
We have attached a video of obtaining a \textit{diamond} in the supplementary materials.

\section{Applications}

\begin{figure}[ht]
  \small
  \centering
  \subfigure[\texttt{Shelter with Farmland}]{
    \includegraphics[height=0.2\linewidth]{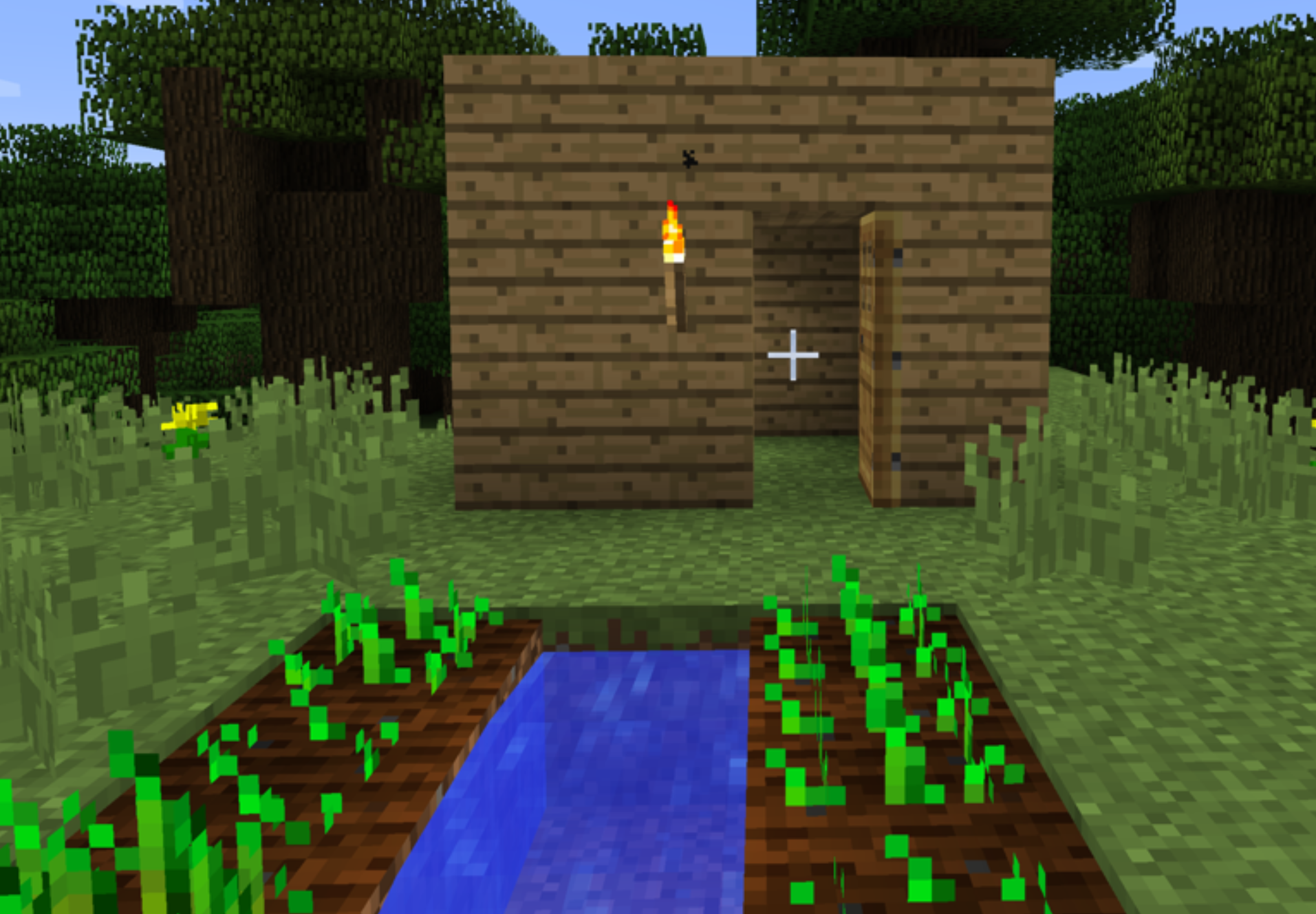}
  }
  \subfigure[\texttt{Iron Golem}]{
    \includegraphics[height=0.2\linewidth]{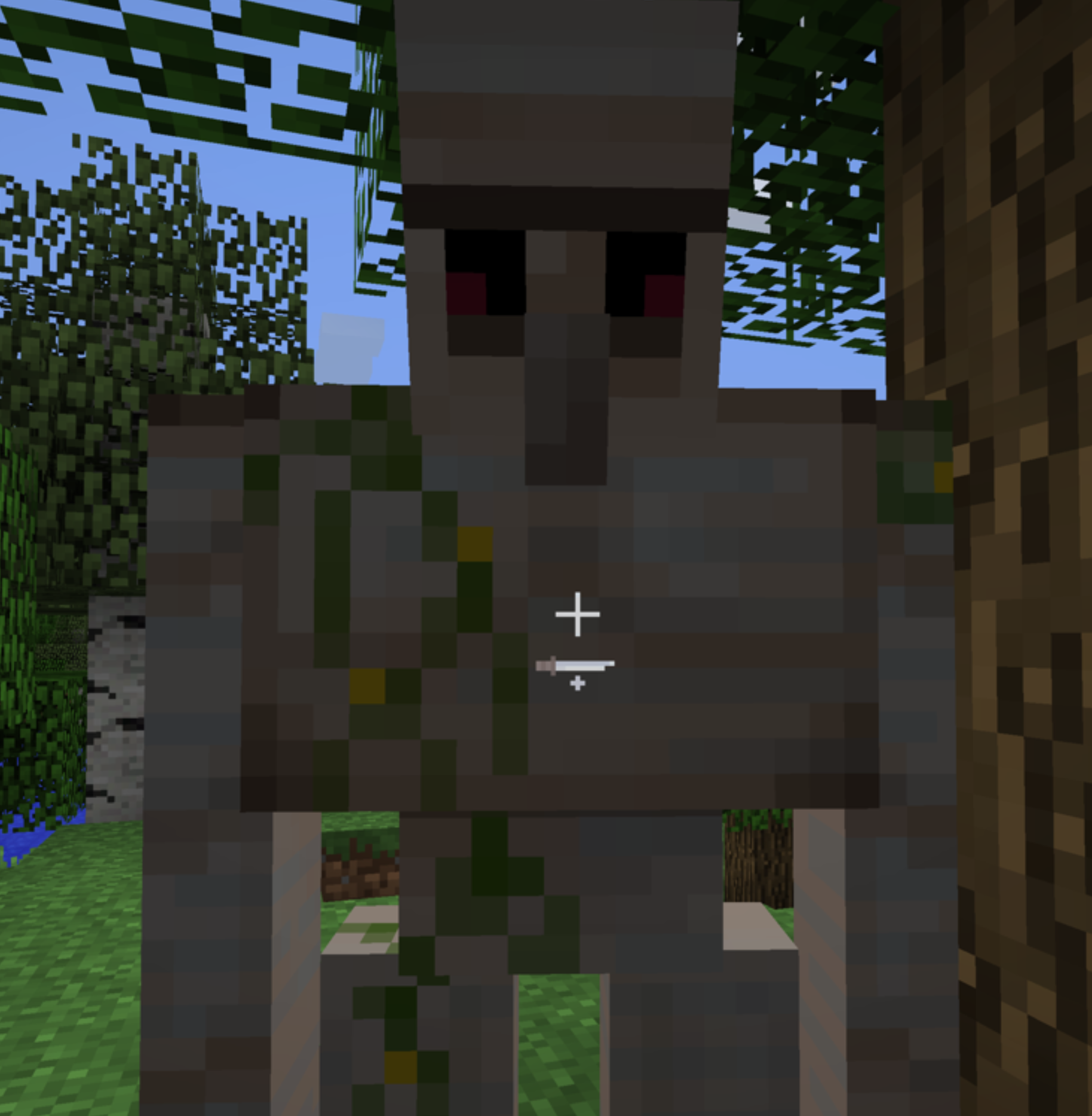}
  }
  \subfigure[\texttt{Redstone Circuit}]{
    \includegraphics[height=0.2\linewidth]{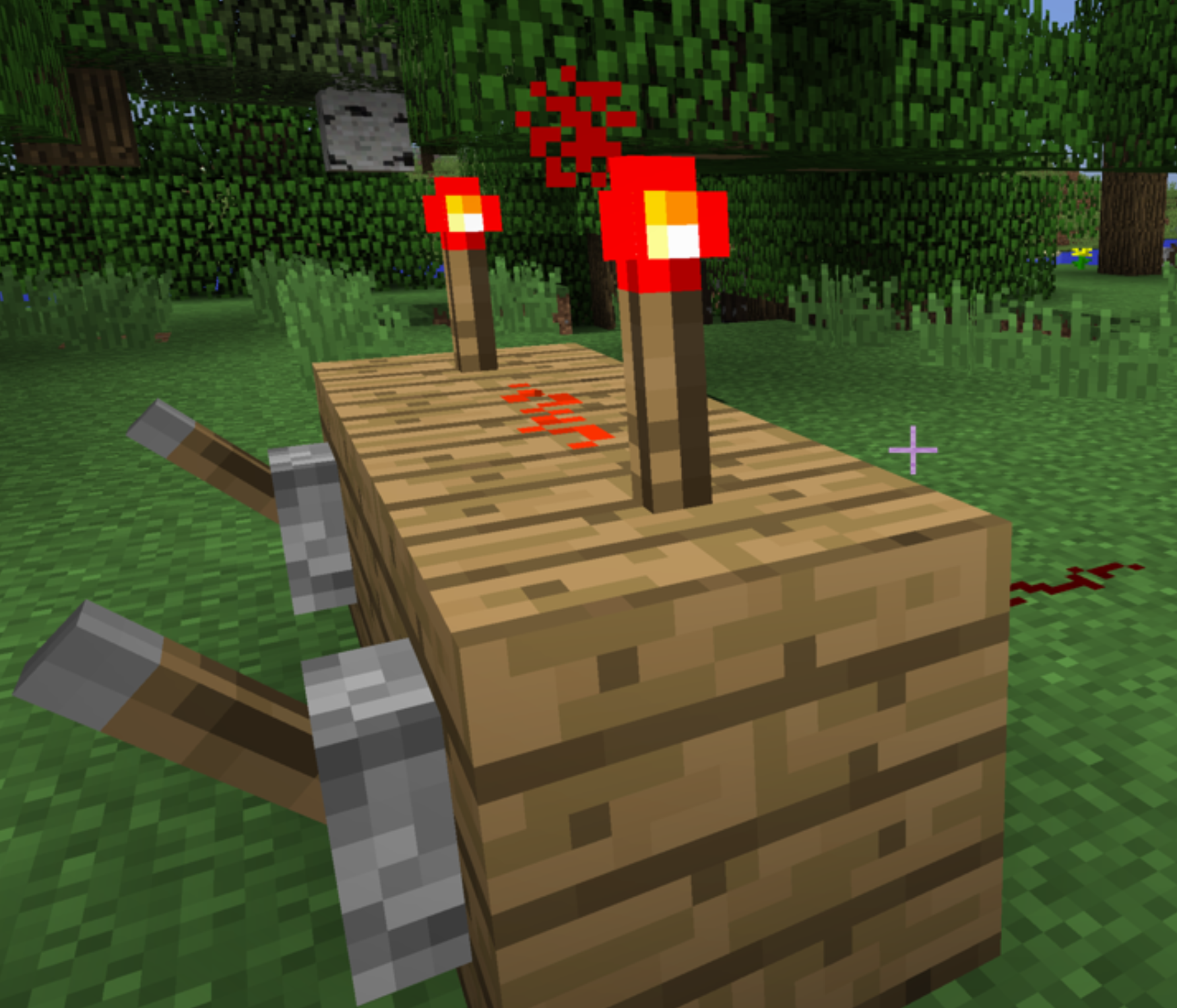}
  }
  \subfigure[\texttt{Nether Portal}]{
    \includegraphics[height=0.2\linewidth]{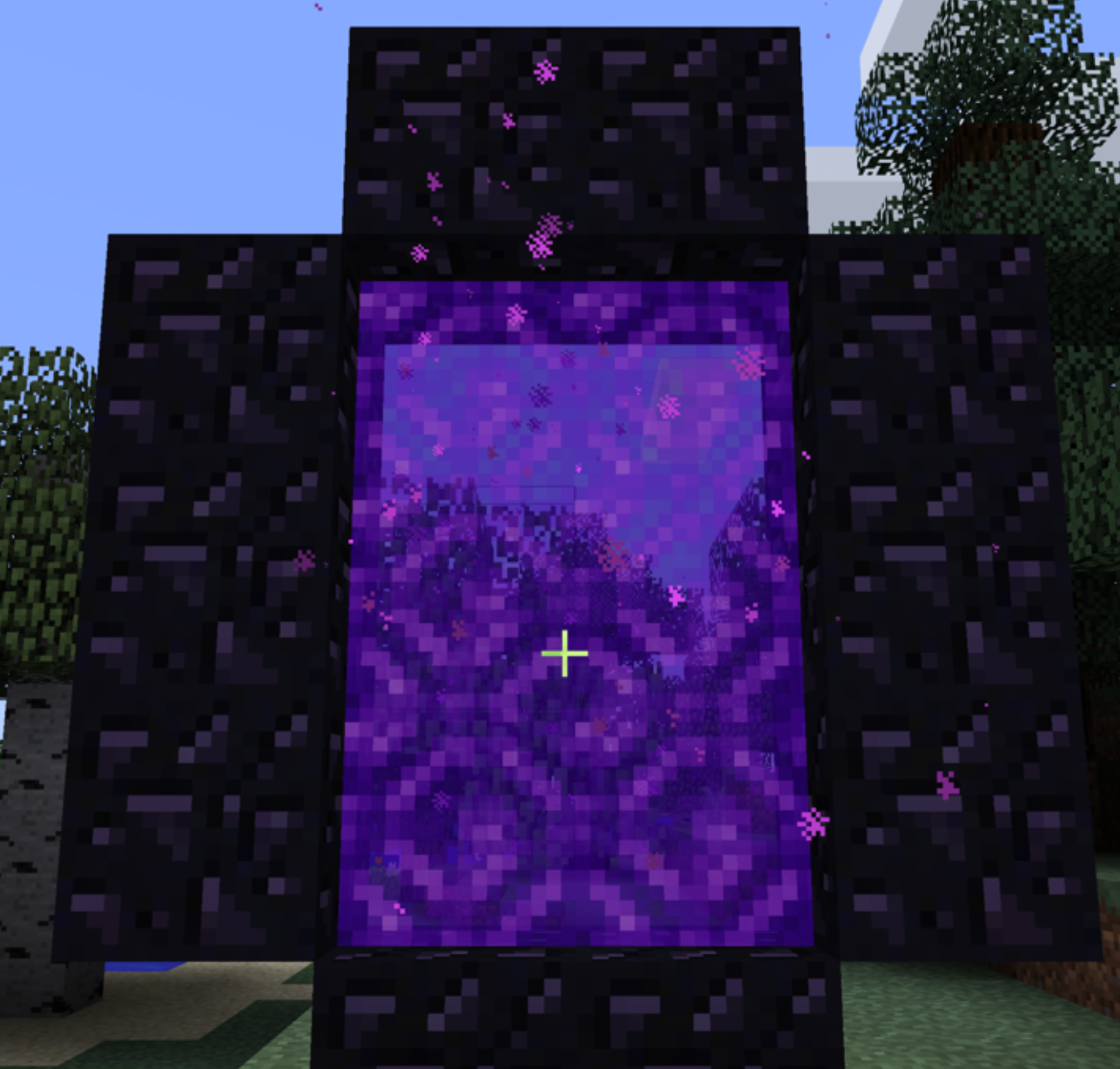}
  }
  \caption{Demonstration of the applications. GITM can construct \texttt{Shelter with Farmland} and \texttt{Iron Golem} for survival, \texttt{Redstone Circuit} for automation equipment, and \texttt{Nether Portal} for the Nether world exploration.}
  \label{fig:application}
\end{figure}

Our proposed GITM makes survival and the nether exploration possible in Minecraft which has never been accomplished by existing agents. To achieve this, our agent builds four necessary items, including \texttt{Shelter with Farmland}, \texttt{Iron Golem}, \texttt{Redstone Circuit}, and \texttt{Nether Portal}, shown in Fig.~\ref{fig:application}. \texttt{Shelter with Farmland} is firstly built to keep the agent from being attacked by monsters at night and provide enough food. \texttt{Iron Golem} can automatically attack monsters to protect the agent and the shelter. \texttt{Redstone Circuit} is the foundation of all automation equipment. \texttt{Nether Portal} is the entrance to the Nether world.

%% file: tables/appendix-all-item-success-rate.tex
\setlength{\tabcolsep}{1pt}
\begin{table}[H]
    \centering
    \small
    \vspace{-0.5em}
    \caption{\textbf{Success rate for all 262 items in the entire Minecraft Overworld Technology Tree.}}
    \label{tab:appendix-all-item}
    \resizebox{\textwidth}{!}{
    \begin{tabular}{lc|lc|lc|lc}
    \toprule
       Item Name  & \makecell{Success \\ Rate} & Item Name & \makecell{Success \\ Rate} & Item Name & \makecell{Success \\ Rate} & Item Name & \makecell{Success \\ Rate} \\
    \midrule
acacia boat                     &   100.0     &   stone sword                     &   100.0     &   gravel                          &   80.0      &   beetroot seeds                  &   40.0      \\
acacia door                     &   100.0     &   stonebrick                      &   100.0     &   iron boots                      &   80.0      &   diamond boots                   &   40.0      \\
acacia fence                    &   100.0     &   sugar                           &   100.0     &   iron trapdoor                   &   80.0      &   diamond helmet                  &   40.0      \\
acacia fence gate               &   100.0     &   tallgrass                       &   100.0     &   leather chestplate              &   80.0      &   golden axe                      &   40.0      \\
acacia stairs                   &   100.0     &   trapdoor                        &   100.0     &   leather leggings                &   80.0      &   golden pickaxe                  &   40.0      \\
beef                            &   100.0     &   wheat                           &   100.0     &   bone block                      &   75.0      &   red mushroom                    &   40.0      \\
birch boat                      &   100.0     &   wheat seeds                     &   100.0     &   bow                             &   75.0      &   red mushroom block              &   40.0      \\
birch door                      &   100.0     &   wooden axe                      &   100.0     &   chest minecart                  &   75.0      &   diamond leggings                &   35.0      \\
birch fence                     &   100.0     &   wooden button                   &   100.0     &   furnace minecart                &   75.0      &   golden boots                    &   35.0      \\
birch fence gate                &   100.0     &   wooden door                     &   100.0     &   hopper                          &   75.0      &   ink sac                         &   35.0      \\
birch stairs                    &   100.0     &   wooden hoe                      &   100.0     &   iron helmet                     &   75.0      &   sticky piston                   &   35.0      \\
boat                            &   100.0     &   wooden pickaxe                  &   100.0     &   minecart                        &   75.0      &   beetroot soup                   &   30.0      \\
bone                            &   100.0     &   wooden pressure plate           &   100.0     &   mossy cobblestone               &   75.0      &   golden helmet                   &   30.0      \\
bone meal                       &   100.0     &   wooden shovel                   &   100.0     &   vine                            &   75.0      &   lead                            &   30.0      \\
bowl                            &   100.0     &   wooden slab                     &   100.0     &   waterlily                       &   75.0      &   map                             &   30.0      \\
chest                           &   100.0     &   wooden sword                    &   100.0     &   banner                          &   70.0      &   mushroom stew                   &   30.0      \\
chicken                         &   100.0     &   yellow flower                   &   100.0     &   brick                           &   70.0      &   brick stairs                    &   25.0      \\
cobblestone                     &   100.0     &   armor stand                     &   95.0      &   clay ball                       &   70.0      &   cactus                          &   25.0      \\
cobblestone wall                &   100.0     &   book                            &   95.0      &   diamond                         &   70.0      &   cake                            &   25.0      \\
cooked beef                     &   100.0     &   bread                           &   95.0      &   diamond shovel                  &   70.0      &   clock                           &   25.0      \\
cooked chicken                  &   100.0     &   coal                            &   95.0      &   dropper                         &   70.0      &   cooked rabbit                   &   25.0      \\
cooked mutton                   &   100.0     &   fireworks                       &   95.0      &   feather                         &   70.0      &   diamond chestplate              &   25.0      \\
cooked porkchop                 &   100.0     &   gunpowder                       &   95.0      &   iron bars                       &   70.0      &   obsidian                        &   25.0      \\
crafting table                  &   100.0     &   iron ingot                      &   95.0      &   iron door                       &   70.0      &   rabbit                          &   25.0      \\
dark oak boat                   &   100.0     &   iron nugget                     &   95.0      &   jukebox                         &   70.0      &   rabbit hide                     &   25.0      \\
dark oak door                   &   100.0     &   iron ore                        &   95.0      &   lapis lazuli                    &   70.0      &   deadbush                        &   20.0      \\
dark oak fence                  &   100.0     &   iron shovel                     &   95.0      &   noteblock                       &   70.0      &   golden leggings                 &   20.0      \\
dark oak fence gate             &   100.0     &   item frame                      &   95.0      &   piston                          &   70.0      &   golden rail                     &   20.0      \\
dark oak stairs                 &   100.0     &   leather                         &   95.0      &   rail                            &   70.0      &   lapis block                     &   20.0      \\
dirt                            &   100.0     &   rotten flesh                    &   95.0      &   redstone                        &   70.0      &   writable book                   &   20.0      \\
double plant                    &   100.0     &   shield                          &   95.0      &   redstone torch                  &   70.0      &   baked potato                    &   15.0      \\
fence                           &   100.0     &   spider eye                      &   95.0      &   cauldron                        &   65.0      &   carrot                          &   15.0      \\
fence gate                      &   100.0     &   stone slab                      &   95.0      &   diamond hoe                     &   65.0      &   diamond block                   &   15.0      \\
furnace                         &   100.0     &   torch                           &   95.0      &   diamond sword                   &   65.0      &   emerald block                   &   15.0      \\
glass                           &   100.0     &   trapped chest                   &   95.0      &   emerald                         &   65.0      &   golden chestplate               &   15.0      \\
glass bottle                    &   100.0     &   tripwire hook                   &   95.0      &   iron leggings                   &   65.0      &   potato                          &   15.0      \\
glass pane                      &   100.0     &   carpet                          &   90.0      &   tnt                             &   65.0      &   pumpkin                         &   15.0      \\
ladder                          &   100.0     &   coal block                      &   90.0      &   arrow                           &   60.0      &   pumpkin seeds                   &   15.0      \\
lever                           &   100.0     &   grass                           &   90.0      &   compass                         &   60.0      &   carrot on a stick               &   10.0      \\
log                             &   100.0     &   heavy weighted pressure plate   &   90.0      &   flower pot                      &   60.0      &   jungle boat                     &   10.0      \\
mutton                          &   100.0     &   iron hoe                        &   90.0      &   iron chestplate                 &   60.0      &   jungle door                     &   10.0      \\
oak stairs                      &   100.0     &   iron sword                      &   90.0      &   brick block                     &   55.0      &   jungle fence                    &   10.0      \\
paper                           &   100.0     &   leather boots                   &   90.0      &   clay                            &   55.0      &   jungle fence gate               &   10.0      \\
planks                          &   100.0     &   leather helmet                  &   90.0      &   dispenser                       &   55.0      &   jungle stairs                   &   10.0      \\
porkchop                        &   100.0     &   leaves                          &   90.0      &   gold ingot                      &   55.0      &   lit pumpkin                     &   10.0      \\
red flower                      &   100.0     &   painting                        &   90.0      &   gold nugget                     &   55.0      &   melon                           &   10.0      \\
reeds                           &   100.0     &   shears                          &   90.0      &   gold ore                        &   55.0      &   melon block                     &   10.0      \\
sand                            &   100.0     &   snow                            &   90.0      &   golden shovel                   &   55.0      &   melon seeds                     &   10.0      \\
sandstone                       &   100.0     &   snow layer                      &   90.0      &   hardened clay                   &   55.0      &   gold block                      &   8.0       \\
sapling                         &   100.0     &   snowball                        &   90.0      &   hopper minecart                 &   55.0      &   golden carrot                   &   8.0       \\
sign                            &   100.0     &   string                          &   90.0      &   iron block                      &   55.0      &   pumpkin pie                     &   8.0       \\
spruce boat                     &   100.0     &   wool                            &   90.0      &   slime ball                      &   55.0      &   red sandstone                   &   6.0       \\
spruce door                     &   100.0     &   bed                             &   85.0      &   activator rail                  &   50.0      &   red sandstone stairs            &   6.0       \\
spruce fence                    &   100.0     &   brown mushroom                  &   85.0      &   detector rail                   &   50.0      &   speckled melon                  &   6.0       \\
spruce fence gate               &   100.0     &   brown mushroom block            &   85.0      &   diamond axe                     &   50.0      &   stone slab2                     &   6.0       \\
spruce stairs                   &   100.0     &   bucket                          &   85.0      &   diamond pickaxe                 &   50.0      &   anvil                           &   4.0       \\
stick                           &   100.0     &   flint                           &   85.0      &   egg                             &   50.0      &   apple                           &   4.0       \\
stone                           &   100.0     &   hay block                       &   85.0      &   lava bucket                     &   50.0      &   enchanting table                &   4.0       \\
stone axe                       &   100.0     &   iron axe                        &   85.0      &   repeater                        &   50.0      &   enchanted book                  &   3.0       \\
stone brick stairs              &   100.0     &   iron pickaxe                    &   85.0      &   tnt minecart                    &   50.0      &   poisonous potato                &   2.0       \\
stone button                    &   100.0     &   milk bucket                     &   85.0      &   bookshelf                       &   45.0      &   golden apple                    &   1.0       \\
stone hoe                       &   100.0     &   sandstone stairs                &   85.0      &   golden hoe                      &   45.0      &   rabbit foot                     &   1.0       \\
stone pickaxe                   &   100.0     &   water bucket                    &   85.0      &   golden sword                    &   45.0      &   slime                           &   1.0       \\
stone pressure plate            &   100.0     &   fermented spider eye            &   80.0      &   light weighted pressure plate   &   45.0      &   rabbit stew                     &   0.5       \\
stone shovel                    &   100.0     &   fishing rod                     &   80.0      &   redstone block                  &   45.0      &   & \\
stone stairs                    &   100.0     &   flint and steel                 &   80.0      &   beetroot                        &   40.0      &   & \\
    \bottomrule
    \end{tabular}
    }
    \vspace{-1.5em}
\end{table}

%% file: tables/appendix-ablation.tex
\setlength{\tabcolsep}{5pt}
\setlength{\doublerulesep}{2\arrayrulewidth}
\begin{table}[ht]
    \centering
    \small
    \caption{\textbf{Ablation study.} The milestone items from left to right are crafting table \mccraftingtable, wooden pickaxe \mcwoodenpickaxe, stone pickaxe \mcstonepickaxe, iron pickaxe \mcironpickaxe, and diamond \mcdiamond. The success rate is calculated under time limit of 12000 steps (total) and query limit of 30 (each sub-goal). ``Goal Decomp.'' indicates whether to use LLM Decomposer to decompose the goal into sub-goals. ``Goal / Sub-Goal to Structured Actions / Keyboard \& Mouse Mapping'' indicates which method is used for the mapping from goal / sub-goals to structured actions / keyboard \& mouse operations.}
    \vspace{0.5em}
    \label{tab:appendix_ablation}
    \resizebox{\textwidth}{!}{
    \begin{tabular}{cccc|ccccc}
    \toprule
    & \multirow{3}{*}{\makecell{Goal \\ Decomp.}} & \multirow{3}{*}{\makecell{Structured \\ Action}} & \multirow{3}{*}{\makecell{Goal / Sub-Goal to \\ Structured Actions / Keyboard \& Mouse \\ Mapping}} & \multicolumn{5}{c}{Success Rate (\%)}\\
    & & & & \multirow{2}{*}{\mccraftingtable} & \multirow{2}{*}{\mcwoodenpickaxe} & \multirow{2}{*}{\mcstonepickaxe} & \multirow{2}{*}{\mcironpickaxe} & \multirow{2}{*}{\mcdiamond} \\
    \\
    \midrule
     (a) & & & Specialist RL Model (VPT) & 100.0 & 100.0 & 100.0 & 85.0 & 20.0 \\
     (b) & & & Goal-conditioned RL Model (DEPS) & 0.0 & 0.0 & 0.0 & 0.0 & 0.0 \\
     (c) & & & Our LLM Planner & 0.0 & 0.0 & 0.0 & 0.0 & 0.0 \\
    \midrule
     (d) & \checkmark & & Goal-conditioned RL Model (DEPS) & 90.0 & 80.0 & 30.0 & 0.0 & 0.0 \\
     (e) & \checkmark & & Our LLM Planner & 0.0 & 0.0 & 0.0 & 0.0 & 0.0 \\
     (f) & & \checkmark & Our LLM Planner & 57.5 & 32.5 & 5.0 & 0.0 & 0.0\\
     (g) & \checkmark & \checkmark & Our LLM Planner & 100.0 & 100.0 & 100.0 & 95.0 & 67.5 \\
    \bottomrule
    \end{tabular}
    }
\end{table}